\documentclass[10pt,twocolumn,letterpaper]{article}

\usepackage{cvpr}              

\usepackage{arydshln}

\renewcommand{\paragraph}[1]{\vspace{.25em}\noindent\textbf{#1.}}

\newcommand{\methodname}{ViFCLIP\xspace}
\newcommand{\baselinename}{LongViCLIP\xspace}

\definecolor{cvprblue}{rgb}{0.21,0.49,0.74}
\usepackage[pagebackref,breaklinks,colorlinks,allcolors=cvprblue]{hyperref}
\usepackage{booktabs}
\usepackage{multirow}
\usepackage{makecell}
\usepackage{graphicx}
\usepackage{xcolor}     
\usepackage[table]{xcolor}
\usepackage{tabularx}
\usepackage{array}
\newcommand{\topstack}[1]{%
    \begin{tabular}[t]{@{}c@{}}#1\end{tabular}%
}

\definecolor{incrgreen}{RGB}{0,150,70}
\definecolor{decrred}{RGB}{200,30,30}
\definecolor{crossencodergray}{gray}{0.94}

\def\confName{CVPR}
\def\confYear{2027}

\title{ViFCLIP: Video-text Representation Learning\\with Fine-grained Video-text Alignment}

\author{
Hamza Farach\textsuperscript{1,3,4*} \quad
Renaud Vandeghen\textsuperscript{1*} \quad
Fida Mohammad Thoker\textsuperscript{2*} \\
Bernard Ghanem\textsuperscript{2} \quad
Anthony Cioppa\textsuperscript{1} \\[4pt]
\textsuperscript{1}University of Li\`ege \quad
\textsuperscript{2}KAUST \quad
\textsuperscript{3}Keio University \quad
\textsuperscript{4}\'Ecole Centrale de Nantes
}

\begin{document}
\maketitle

\begingroup
\renewcommand{\thefootnote}{\fnsymbol{footnote}}
\footnotetext[1]{Equal contribution.}
\endgroup

\begin{abstract}

Video-text representation learning has shown promising capabilities in many multi-modal applications. Current methods follow the de-facto vision-text alignment during pre-training, allowing a shared representation. However, video-text alignment methods train on coarse-grained text description, resulting in weak representations for detailed, fine-grained descriptions. While some approaches tried to incorporate long captions for learning fine-grained representations, they only rely on a global video-text alignment, failing to effectively capture such fine-grained details. To address this issue, we propose \emph{\methodname}, a video-text model trained with fine-grained video-text alignment. Our model integrates learnable frame-tokens, allowing a fine-grained visual representation beyond the global video representation. We also adopt a subcaption sampling strategy that generates diverse and multi-granular fine-grained textual supervision extracted from detailed captions. Finally, our fine-grained alignment method relies on a token-to-token interaction where each subcaption is aligned with its most similar frame-token. Our model shows SOTA results on many zero-shot text-to-video retrieval datasets against prior video-text dual encoders and MLLM models. Project page: \url{https://uliege-viulab.github.io/vifclip}.

\end{abstract}
\section{Introduction}
\label{sec:intro}

\begin{figure}[t]
    \centering
    \includegraphics[width=0.95\columnwidth]{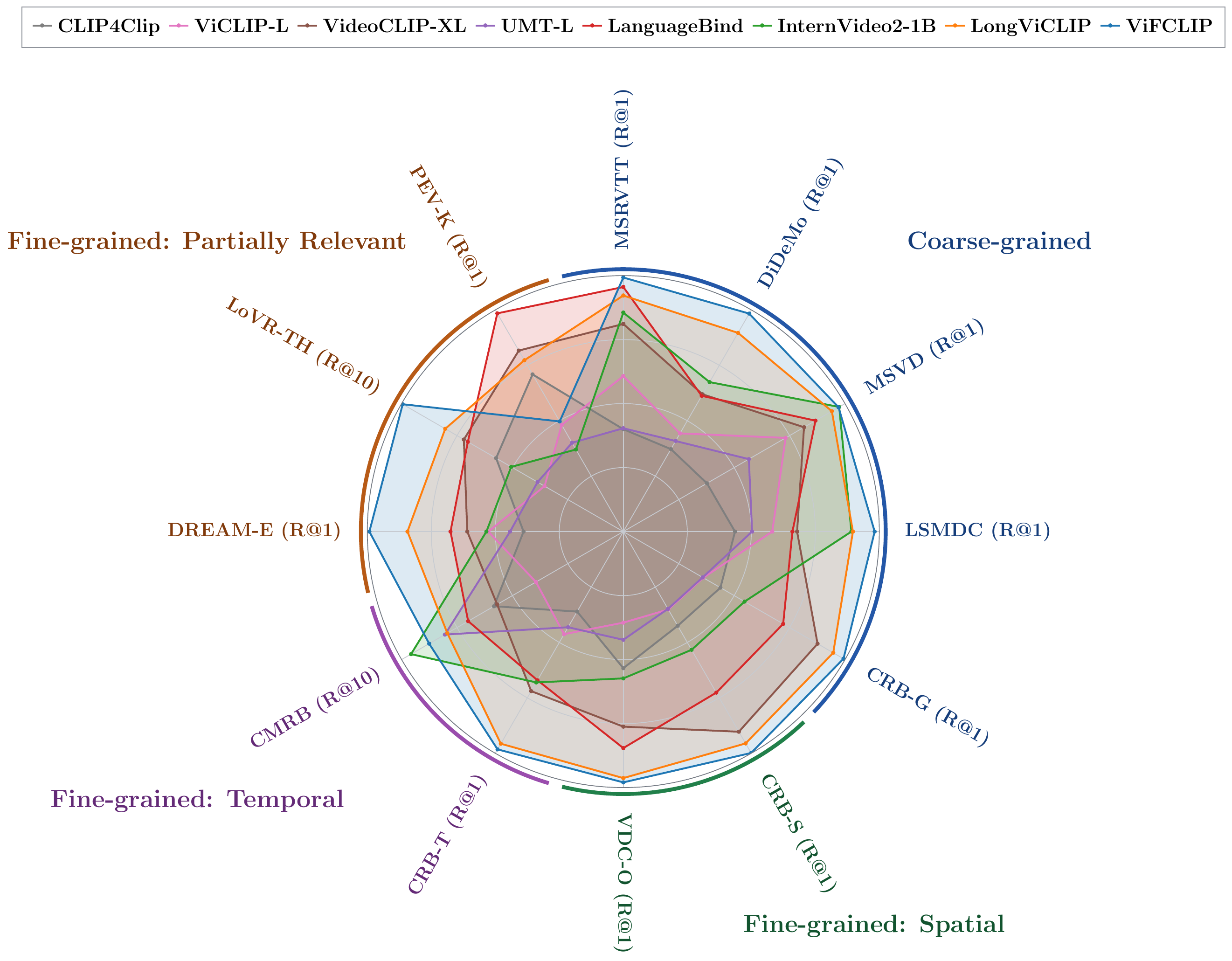}
    \caption{\textbf{\methodname is a strong video-text model.} Against prior methods, our method shows state-of-the-art results on most zero-shot text-to-video retrieval benchmarks.}
    \label{fig:retrieval_radar}
\end{figure}
Visual content can be described at different levels of granularity, from a broad description of a scene to specific objects, attributes, and interactions. Learning representations that capture these different levels of semantics is a central challenge in vision-language pretraining. CLIP~\cite{Radford2021Learning} has demonstrated the effectiveness of contrastive learning for aligning images and natural language in a shared embedding space, enabling strong generalization to unseen visual concepts. Video-text pretraining methods~\cite{Bain2021Frozen,Li2023Unmasked,Luo2022CLIP4Clip,Wang2024InternVid,Wang2024InternVideo2,Doughty2024LocoMotion} further extend contrastive alignment to videos, where visual representations must also account for temporal information. In particular, CLIP-style dual encoders typically represent an entire video with a single embedding, which is aligned with its corresponding textual description. While this global alignment is effective at capturing the overall semantics of a video, it provides limited supervision to distinct individual actions, objects, and interactions occurring in the video~\cite{Xiong2024Structured, Wang2023Unified, Reddy2025VideocolBERT}.

Recent advances in image-text pretraining have explored richer textual supervision to improve the fine-grained understanding of visual content. Methods such as Long-CLIP~\cite{Zhang2024LongCLIP}, LoTLIP~\cite{Wu2024LoTLIP}, and DreamLIP~\cite{Zheng2024DreamLIP} leverage long descriptions or multiple subcaptions to expose models to visual details that are often absent from short captions. Other approaches, including FILIP~\cite{Yao2022FILIP}, FLAIR~\cite{Xiao2025FLAIR}, and $\beta$-CLIP~\cite{Zohra2026betaCLIP}, further investigate fine-grained alignment by establishing correspondences between textual descriptions and localized visual features. However, extending these approaches to video-text pretraining introduces additional challenges. Unlike images, videos contain temporally distributed visual information, where different parts of a description may refer to distinct moments of the sequence. Moreover, a caption may describe only a subset of the observed content, making a noisy alignment between every visual element and its associated text. Although recent video-text methods started exploiting detailed descriptions, multiple captions or fine-grained video-text correspondences~\cite{Thoker2026Advancing-arxiv,Wang2024VideoCLIPXL,Xiong2024Structured}, they do not explicitly associate individual textual details with temporally localized visual representations. Consequently, the additional information provided by richer captions may not be fully exploited when supervision is restricted to a single global video embedding.

In this work, we propose \textbf{\methodname}, a video-text pretraining framework that jointly learns global and fine-grained visual representations through contrastive alignment. Building upon ViCLIP~\cite{Wang2024InternVid}, we leverage captions at different granularities~\cite{Thoker2026Advancing-arxiv} to supervise both global and localized visual representations. Rather than relying solely on global alignment, \methodname establishes fine-grained correspondences between subcaptions and their most relevant visual tokens, without explicit temporal supervision.

\paragraph{Contributions} Our main contributions are summarized as follows:
(1) We introduce \baselinename, a strong video-text pretraining baseline that extends ViCLIP to learn from multiple summary and detailed captions, 
(2) we introduce \methodname, a video-text pretraining framework that jointly learns global and fine-grained representations. Specifically, we enhance the video encoder with learnable register and frame-tokens, temporal attention blocks, and a subcaption sampling strategy generating fine-grained textual descriptions. The contrastive objective then aligns each subcaption with its most relevant visual token through maximum similarity, 
and (3) we demonstrate the effectiveness of \methodname through extensive zero-shot video retrieval experiments, achieving state-of-the-art results on several coarse-grained, fine-grained, and long-context retrieval benchmarks.

\section{Related Work}
\label{sec:related}

\noindent\textbf{Video-text pretraining.}
Video-text pretraining aims to learn representations that capture the relationship between video content and language.
Early work explored different strategies to learn joint video-text representations. ClipBERT~\cite{Lei2021Less} uses sparse frame sampling for end-to-end pretraining, while Frozen in Time~\cite{Bain2021Frozen} trains a shared space-time encoder on both images and videos. UMT~\cite{Li2023Unmasked} improves training efficiency by aligning visible video tokens with features from an unmasked CLIP teacher. Other works like InternVideo~\cite{Wang2022InternVideo-arxiv} and InternVideo2~\cite{Wang2024InternVideo2} combine masked video modeling~\cite{Tong2022VideoMAE,Thoker2025SMILE,Vandeghen2026TrackMAE} with video-language alignment at scale.

Another line of work adapts CLIP~\cite{Radford2021Learning} to video. CLIP4Clip~\cite{Luo2022CLIP4Clip} independently encodes each frame using the CLIP image encoder and aggregates their features for retrieval. ViCLIP~\cite{Wang2024InternVid} initializes both encoders from CLIP and adds spatiotemporal attention to the visual encoder. Following CLIP, ViCLIP retains the dual-encoder design and aligns a global video embedding with its caption through contrastive learning. We build on ViCLIP to learn from multiple short and detailed captions per video. 

\noindent\textbf{Learning from detailed captions.} Long and detailed captions provide supervision for objects, attributes, and relationships that short descriptions often omit
\cite{Chen2024ShareGPT4V,Zhang2024LongCLIP,Wu2024LoTLIP, Zheng2024DreamLIP,Asokan2025FineLIP,Xiao2025FLAIR, Xie2025FGCLIP,Zohra2026betaCLIP}. Image-text methods enrich this supervision by generating detailed captions for existing images, synthesizing image-text pairs, or rewriting existing captions \cite{Chen2024ShareGPT4V,Hammoud2024SynthCLIP-arxiv,Fan2023Improving}. Long-CLIP~\cite{Zhang2024LongCLIP} and LoTLIP~\cite{Wu2024LoTLIP} study how to learn from long descriptions while retaining performance on short text. Fine-grained methods align text with local visual features through patch grouping \cite{Yao2022FILIP,Asokan2025FineLIP,Bica2024Improving}, region-level supervision \cite{Zhong2022RegionCLIP,Xie2025FGCLIP,Choi2025GOAL,Kawamura2026PowerCLIP}, or text-conditioned pooling and feature selection \cite{Lavoie2024Modeling,Xiao2025FLAIR,Zohra2026betaCLIP,Xie2025SmartCLIP}. In particular,  DreamLIP~\cite{Zheng2024DreamLIP} samples subcaptions and aligns them with groups of patches. Llip~\cite{Lavoie2024Modeling} conditions the aggregation of learned visual tokens on text, while FLAIR \cite{Xiao2025FLAIR} and $\beta$-CLIP \cite{Zohra2026betaCLIP} use text to directly pool relevant image patches. Our proposed \methodname framework extends this setting to video, where multiple short and detailed captions provide supervision for global and fine-grained alignment.

For video, ShareGPT4Video and Panda-70M use generative models to caption video clips at scale~\cite{Chen2024ShareGPT4Video,Chen2024Panda70M}. Cap4Video~\cite{Wu2023Cap4Video} uses generated captions as additional positives during training and incorporates them into retrieval. VideoCLIP-XL~\cite{Wang2024VideoCLIPXL} extends ViCLIP to learn from both short and long captions, and applies PCA on the video features for coarse-grained alignment, while introducing new learning tasks to better model long captions.~\cite{Xiong2024Structured} extends global video-text alignment with spatial grounding and temporal grouping to capture fine-grained region-object correspondences. However, it focuses on aligning individual objects rather than richer descriptions of video content. Recently, MVC~\cite{Thoker2026Advancing-arxiv} generates multiple short and detailed descriptions per video and uses them to align a masked video-modeling encoder with a text encoder based on~\cite{Li2023Unmasked}. Building on this textual supervision, we adapt ViCLIP-style dual encoders to learn from multiple short and long captions for the same video. We then refine this alignment by matching subcaptions drawn from the detailed descriptions to frame-level tokens to further improve fine-grained alignment.
\section{Methodology}
\label{sec:methodology}
\begin{figure*}[t]
    \centering
    \includegraphics[width=\linewidth]{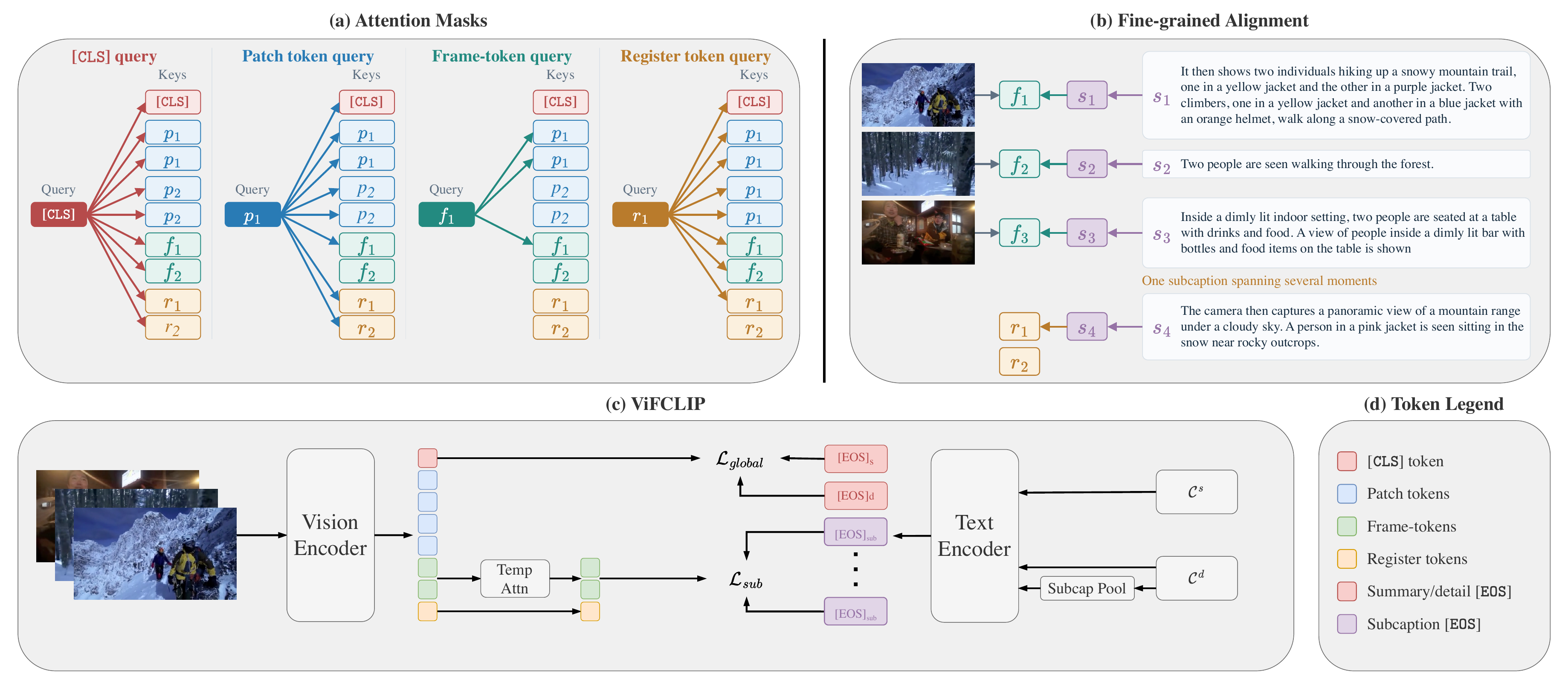}
    \caption{\textbf{Overview of \methodname.}
    \textbf{(a)} Interaction between different tokens in the vision encoder.
    \textbf{(b)} Subcaption generation and MaxSim operation for the fine-grained alignment.
    \textbf{(c)} The vision encoder processes sampled frames of a video to produce global [\texttt{CLS}], local frame-tokens and register tokens. The text encoder processes one summary caption, one detailed caption, and $K$ subcaptions to produce their relative [\texttt{EOS}] representations. Global vision and text representations are aligned together, while registers and frame-tokens are aligned to subcaptions in a single joint vision-text embedding space.
    \textbf{(d)} Legend relative to the different tokens.
    }
    \label{fig:method}
\end{figure*}

\subsection{Preliminaries: ViCLIP}
\label{sec:prelim_viclip}
ViCLIP~\cite{Wang2024InternVid} extends CLIP~\cite{Radford2021Learning} to learn a joint video-text embedding space using a large collection of video-text pairs  $(V,T)$.
The model consists of two encoders: a vision encoder built with a spatiotemporal ViT~\cite{Dosovitskiy2021AnImage} and a text encoder, both initialized from CLIP~\cite{Radford2021Learning}. The input video $V$ is subsampled in frames that are partitioned into a set of patch tokens. These patch tokens receive both spatial and temporal positional embeddings and are processed by the vision encoder which outputs a \texttt{[CLS]} token to represent the video. The caption $T$ is tokenized and processed by the text encoder which outputs a \texttt{[EOS]} token to represent the text. The two encoders use separate projection heads to map their respective output tokens into a shared embedding space, yielding the video, $v_{\mathrm{cls}}$, and text, $t_{\mathrm{eos}}$, embeddings. Given a batch of $B$ video-text pairs $\{(V_i,T_i)\}_{i=1}^{B}$, ViCLIP computes the cosine similarity $s_{ij}=\mathrm{sim}(v_{\mathrm{cls}}^i,t_{\mathrm{eos}}^j)$ between every video and text in the batch, \ie $\forall\, i,j \in \{1,\dots,B\}$. The objective function $\mathcal{L}_{\mathrm{CLIP}}$ is based on the InfoNCE loss~\cite{VanDenOord2018Representation-arxiv} and aims to maximize the similarity between matching video-text pairs while minimizing the similarity between non-matching pairs:
\begin{equation}
\label{eq:infonce}
\begin{aligned}
\mathcal{L}_{\mathrm{CLIP}}
= -\frac{1}{2B}\sum_{i=1}^{B}
\Bigg[
&\overbrace{
\log\frac{\exp(s_{ii}/\tau)}
{\sum_{j=1}^{B}\exp(s_{ij}/\tau)}
}^{\text{video-to-text}}
\\
+&\overbrace{
\log\frac{\exp(s_{ii}/\tau)}
{\sum_{j=1}^{B}\exp(s_{ji}/\tau)}
}^{\text{text-to-video}}
\Bigg],
\end{aligned}
\end{equation}
where \(\tau\) is a learnable temperature parameter.

\subsection{Adapting ViCLIP to Multiple Captions}
\label{sec:viclip_multiple_captions}
We first adapt ViCLIP to learn from multiple short and long captions per video. For image-text pretraining, Long-CLIP~\cite{Zhang2024LongCLIP} uses short and long captions to supervise different levels of detail. DreamLIP~\cite{Zheng2024DreamLIP} uses several descriptions of the same image as positives for global contrastive learning. Since these two ideas are complementary for videos, training on both types of captions allows the same video to be matched to a short and a long description. To do so, we use the captions from~\cite{Thoker2026Advancing-arxiv}, which provide multiple summary and detailed descriptions for the videos present in InternVid-10M-FLT~\cite{Wang2024InternVid} dataset. Each video $V_i$ is associated with a set of summary captions $\mathcal{C}_i^s$ and a set of detailed captions $\mathcal{C}_i^d$, respectively. At each training step, we randomly sample one caption from each set:
\begin{equation}
c_i^s \sim \mathcal{C}_i^s, \qquad c_i^d \sim \mathcal{C}_i^d,
\end{equation}
that are processed by the same text encoder, and the video is encoded once. The training objective is the sum of two losses using \cref{eq:infonce}:
\begin{equation}
\label{eq:global_caption_alignment}
\mathcal{L}_{\mathrm{global}} = \mathcal{L}_{\mathrm{CLIP}}^s + \mathcal{L}_{\mathrm{CLIP}}^d.
\end{equation}
Training on this global-alignment provides a baseline model, called \emph{\baselinename}, that jointly trains the vision and text encoders on captions of multiple granularities. To account for longer captions, we extend the context length of the text encoder, following~\cite{Zhang2024LongCLIP}.

However, the global objective in \emph{\baselinename} still compares each detailed caption with a single video embedding, while a detailed caption can describe several actions or interactions that occur at different times. Hence, all of these details contribute to the same video-text similarity score. In image-text learning, FLAIR~\cite{Xiao2025FLAIR} and $\beta$-CLIP~\cite{Zohra2026betaCLIP} accommodate for this issue by using parts of a caption to guide the pooling of image patches. Since in video, the visual evidence for different parts of a caption may also appear in different frames, an additional objective that lets different parts of a caption supervise different frames would enhance the fine-grained alignment between video and text data.

\subsection{Fine-Grained Video-Text Alignment}
\label{sec:fine-grained}

We build \methodname on \baselinename by adding a contrastive objective between \emph{subcaptions} and \emph{frame-level} video features. We represent each sampled frame with learnable \emph{frame-tokens} and construct \emph{subcaptions} from the sentences present in the detailed caption pool. Each subcaption is matched to the frame-token with the highest similarity. The overall framework is shown in~\cref{fig:method}.

\paragraph{Frame-token \& Attention Mask}
The \texttt{[CLS]} token aggregates information across all frames, limiting its ability to preserve local information across spatial and temporal dimensions. To address this issue, we add learnable frame-tokens to each frames to retain local features. Each frame-token receives the temporal positional embedding of its associated frame. An attention mask restricts each frame-token to attend directly to itself and the visible patches of its own frame. The \texttt{[CLS]} and patch tokens can attend to all tokens. The different interactions between tokens are depicted in~\cref{fig:method} (a). We apply at the output of the vision encoder temporal attention blocks to let the frame tokens attend to each other. We project their resulting embeddings with the same projection head as the \texttt{[CLS]} token.

\paragraph{Subcaption Sampling}
A detailed caption often contains several sentences that describe different actions, objects, or parts of the scene. For each video $V_i$, we split all captions in $\mathcal{C}_i^d$, in single sentences, creating a pool of sentences. To form a subcaption, we randomly select $n \in \{1,\ldots,N\}$ sentences from this pool and concatenate them, with $N$ being the maximum number of sentences. We repeat this process $K$ times. A subcaption can contain a single detail or combine several details, including sentences from different captions of the same video. As for the summary and detailed captions, we encode each subcaption with the same text encoder and denote its projected \texttt{[EOS]} embedding by $t_{i,k}^{\mathrm{sub}}$, where $k \in \{1,\ldots,K\}$.

\paragraph{Fine-grained Alignment}
During training, the frame-token that best matches a subcaption is unknown. Hence, we use the maximum cosine similarity operator (MaxSim) to select the frame-token with the highest similarity with respect to a subcaption. We also introduce learnable register tokens~\cite{Darcet2024Vision} to capture complementary visual information that cannot be assigned to a specific frame-token, such as global context or background content. The registers have no temporal positional embeddings and can attend to all visual tokens.

Let $\mathcal{U}_i$ be the set of projected frame-tokens and registers for video $V_i$, the similarity between $V_i$ and the $k$-th subcaption of video $V_j$ is:

\begin{equation}
\label{eq:maxsim}
s_{i,j,k}^{\mathrm{sub}}
=
\max_{u \in \mathcal{U}_i}
\mathrm{sim}\left(u,t_{j,k}^{\mathrm{sub}}\right).
\end{equation}
Since this process is repeated $K$ times, this lets different subcaptions of the same video match different tokens. Several subcaptions can also match the same token when they
describe the same frame or scene context. This process is highlighted in~\cref{fig:method} (b). At inference, we compute the fine-grained similarity as the MaxSim operator between the text query and the frame-tokens, excluding register tokens. We then average this score with the global video-text similarity to obtain the final retrieval score.

\paragraph{Contrastive objectives}
For a batch of $B$ videos with $K$ subcaptions per video, we use the fine-grained similarity in~\Cref{eq:maxsim} for both contrastive directions. In the video-to-text direction, each video is compared against all $BK$ subcaptions, with its $K$ associated subcaptions as positives. The loss is then averaged over its $K$ positive subcaptions. In the text-to-video direction, each subcaption is compared with all $B$ videos in the batch, with its corresponding video as the only positive. The video-to-text and text-to-video losses are given by:
\begin{align}
\mathcal{L}_{\mathrm{v2sub}}
&=
-\frac{1}{BK}\sum_{i=1}^{B}\sum_{k=1}^{K}
\log
\frac{\exp(s_{i,i,k}^{\mathrm{sub}}/\tau)}
{\sum_{j=1}^{B}\sum_{m=1}^{K}
 \exp(s_{i,j,m}^{\mathrm{sub}}/\tau)},\\
\mathcal{L}_{\mathrm{sub2v}}
&=
-\frac{1}{BK}\sum_{j=1}^{B}\sum_{k=1}^{K}
\log
\frac{\exp(s_{j,j,k}^{\mathrm{sub}}/\tau)}
{\sum_{i=1}^{B}\exp(s_{i,j,k}^{\mathrm{sub}}/\tau)}.
\end{align}
We average the two terms to obtain $\mathcal{L}_{\mathrm{sub}}$ and add it to the global objective in~\Cref{eq:global_caption_alignment}:
\begin{equation}
\label{eq:video_fa_objective}
\mathcal{L}_{\mathrm{total}}
=
\mathcal{L}_{\mathrm{global}}
+
\mathcal{L}_{\mathrm{sub}}.
\end{equation}
Combining all these elements results in our final proposed framework, \emph{\methodname} as shown in~\cref{fig:method} (c).
\section{Experiments}
\label{sec:experiments}

\subsection{Implementation Details}
\label{sec:implementation}
We train 2 model sizes of ViT~\cite{Dosovitskiy2021AnImage}, ViT-B/16 and ViT-L/14, both initialized from CLIP~\cite{Radford2021Learning} weights. The text encoder also follows the original CLIP architectures, with a change of text length to $248$. For the main results, all models are trained on the 10M videos of InternVid-10M-FLT~\cite{Wang2024InternVid} with the multi-view captions introduced in MVC~\cite{Thoker2026Advancing-arxiv}. In particular, each video has $4$ short and $4$ detailed captions per video.  For the \baselinename baseline, we sample from all short and detailed captions. For \methodname,  we construct the bag of sentences using the $4$ detailed captions and define $K=12$ subcaptions created from a maximum of $N=3$ sentences. For each video, we sample $8$ frames uniformly, downsampled to $224\times224$ and masked at $10\%$, that we globally align with one summary and one detailed caption. For the fine-grained alignment, we process one frame-token per frame and use a two-layer temporal attention block, together with two register tokens, to compute the MaxSim operation. The frame-tokens and the register tokens are randomly initialized. We optimize the model with AdamW~\cite{Loshchilov2019Decoupled} with a global batch size fixed to $4096$ and a learning rate warmup during $0.5$ epoch which then decays with a cosine schedule over $10$ epochs. Training uses mixed precision with \texttt{bf16} autocast. All hyperparameter details are listed in the Appendix.

\subsection{Zero-shot Text-to-Video Retrieval}
\label{sec:retrieval}

\paragraph{Datasets and Evaluation} Following prior works on video-text pretraining~\cite{Wang2024InternVid,Wang2024VideoCLIPXL,Luo2022CLIP4Clip,Li2023Unmasked,Wang2024InternVideo2} we evaluate our method on zero-shot text-to-video (T2V) retrieval on standard datasets: MSR-VTT~\cite{Xu2016MSRVTT}, DiDeMo~\cite{Hendricks2017Localizing}, MSVD~\cite{Chen2011Collecting}, ActivityNet Captions~\cite{Caba2015ActivityNet}, LSMDC~\cite{Rohrbach2015ADataset} and Carebench ~\cite{Xu2026CaReBench} (CRB-G) requiring coarse-grained alignment between video and text data. To demonstrate fine-grained and long-text video retrieval capability, we further evaluate on the T2V settings from the Universal Video Retrieval Benchmark (UVRB) ~\cite{Guo2025Towards-arxiv} based on CaReBench~\cite{Xu2026CaReBench}, DREAM-1K~\cite{Wang2024Tarsier-arxiv}, LoVR~\cite{Liang2026LoVR}, and VDC~\cite{Chai2025AuroraCap} datasets.

We compare our \emph{\baselinename} and \emph{\methodname} with SOTA video-text pretraining methods, including CLIP4Clip~\cite{Luo2022CLIP4Clip}, ViCLIP~\cite{Wang2024InternVid}, VideoCLIP-XL~\cite{Wang2024VideoCLIPXL}, UMT~\cite{Li2023Unmasked}, LanguageBind~\cite{Zhu2024LanguageBind}, and InternVideo2~\cite{Wang2024InternVideo2}. Furthermore, to show a broader comparison, we also evaluate multiple MLLM-based embedding models like GME~\cite{Zhang2025Bridging}, Unite~\cite{Kong2025Modality-arxiv}, BGE-VL~\cite{Zhou2025MegaPairs}, VLM2Vec-V2~\cite{Meng2026VLM2VecV2} and GVE~\cite{Guo2025Towards-arxiv}.
All models are evaluated under the same zero-shot retrieval protocol~\cite{Guo2025Towards-arxiv}, using the MTEB framework~\cite{Muennighoff2023MTEB}, and use publicly released checkpoints. Details about models, datasets, and evaluation protocol are reported in the Appendix.

\begin{table}[t]
    \centering
    \caption{\textbf{Coarse-grained zero-shot text-to-video retrieval.}
    We evaluate our methods against state-of-the-art methods on coarse-grained zero-shot retrieval benchmarks. We report text-to-video R@1. Bold denotes the best and underline the second-best result. All models are evaluated with the same protocol.
    }
    \label{tab:retrieval_standard}
    \scriptsize
    \setlength{\tabcolsep}{1.5pt}
    \renewcommand{\arraystretch}{1.0}
    \begin{tabular*}{\columnwidth}{@{\extracolsep{\fill}}cl*{6}{c}@{}}
        \toprule
        & Method & MSRVTT & DiDeMo & ANet & MSVD & LSMDC & CRB-G \\
        & & R@1 & R@1 & R@1 & R@1 & R@1 & R@1 \\
        \midrule

        \addlinespace[4pt]
        \multirow{5}{*}{\rotatebox[origin=c]{90}{\textit{MLLM-based}}}
        & GME-2B & 37.9 & 30.2 & 35.1 & 49.6 & 19.8 & 73.2 \\
        & UNITE-2B & 43.8 & 37.3 & 43.1 & 50.8 & 22.3 & 79.4 \\
        & VLM2Vec-V2 & 29.0 & 30.9 & 22.0 & 39.1 & 16.3 & 80.6 \\
        & BGE-VL & 40.0 & 37.0 & 35.3 & 49.9 & 21.4 & 86.5 \\
        & GVE-3B & 43.0 & 38.9 & 41.9 & 50.5 & 19.4 & 85.0 \\

        \midrule

        \multirow{6}{*}{\rotatebox[origin=c]{90}{\textit{CLIP-style}}}
        & CLIP4Clip & 32.8 & 29.7 & 25.5 & 40.3 & 16.7 & 50.7 \\
        & ViCLIP-L & 38.4 & 32.8 & 25.8 & 51.0 & 19.1 & 45.0 \\
        & VideoCLIP-XL & 43.9 & 40.4 & 47.2 & 53.5 & 20.7 & 82.9 \\
        & UMT-L & 32.9 & 31.3 & 24.0 & 46.0 & 17.8 & 44.8 \\
        & LanguageBind & \underline{47.8} & 40.1 & 44.8 & 55.0 & 20.4 & 71.5 \\
        & InternVideo2-1B & 45.1 & 42.8 & 45.6 & \textbf{58.2} & 24.2 & 58.7 \\

        \midrule

        \multirow{4}{*}{\rotatebox[origin=c]{90}{\textit{\textbf{Ours}}}}
        & \textbf{\baselinename-B} & 43.3 & 50.2 & 48.5 & 52.9 & 19.3 & 86.6 \\
        & \textbf{\methodname-B} & 43.9 & \underline{53.5} & 49.5 & 55.1 & 21.2 & \underline{88.6} \\
        & \textbf{\baselinename-L} & 46.9 & 52.5 & \textbf{55.2} & 57.2 & \underline{24.3} & 88.1 \\
        & \textbf{\methodname-L} & \textbf{48.8} & \textbf{56.3} & \underline{55.1} & \underline{58.1} & \textbf{25.7} & \textbf{91.5} \\
        \bottomrule
    \end{tabular*}
\end{table}

\paragraph{Coarse-grained results}
\Cref{tab:retrieval_standard} reports the results for coarse-grained zero-shot text-to-video R@1. First, we observe that our \mbox{\baselinename-L} improves on all six benchmarks compared to  ViCLIP-L which is trained on the same 10M videos with single short captions. In particular, \mbox{\baselinename-L} improves DiDeMo (32.8\% to 52.5\%) and ActivityNet Captions (25.8\% to 55.2\%), establishing a strong baseline with multiple short and detailed captions. By adding our fine-grained matching, \mbox{\methodname} improves over the \mbox{\baselinename} baseline on all six benchmarks with ViT-B and five of the six with ViT-L. On DiDeMo, R@1 increases by 3.3 and 3.8 points, respectively, while \mbox{\methodname-L} also gains 3.4 points on CRB-G. 

Among the evaluated dual encoders and MLLM-based embedding models, \mbox{\methodname-L} achieves the highest R@1 on four benchmarks and ranks second on the remaining two. On MSR-VTT, it reaches 48.8\%, 1.0 point above LanguageBind, and on MSVD it reaches 58.1\%, within 0.1 point of InternVideo2-1B. On DiDeMo and LSMDC, it raises the best scores among prior methods from 42.8\% to 56.3\% and from 24.2\% to 25.7\%, respectively, with InternVideo2-1B providing both reference scores. On CRB-G, it raises the best prior score from 86.5\% with BGE-VL to 91.5\%. Even with a ViT-B backbone, \mbox{\methodname-B} is competitive and outperforms all prior models on DiDeMo, ActivityNet Captions, and CRB-G, and matches VideoCLIP-XL on MSR-VTT (43.9\%). Overall, our fine-grained approach preserves strong coarse-grained retrieval performance while providing further gains across both model sizes.

\begin{table*}[t]
    \centering
    \caption{\textbf{Fine-grained and long-context zero-shot text-to-video retrieval.}
    We evaluate our methods against state-of-the-art methods on fine-grained and long-context zero-shot retrieval benchmarks. We report text-to-video R@1 or R@10. Bold denotes the best and underline second-best result. All models are evaluated with the same protocol.
    }
    \label{tab:retrieval_finegrained}
    \scriptsize
    \setlength{\tabcolsep}{1.5pt}
    \renewcommand{\arraystretch}{1.0}
    \begin{tabular*}{\textwidth}{@{\extracolsep{\fill}}cl*{9}{c}@{}}
        \toprule
         & & \multicolumn{7}{c}{Fine-grained} & \multicolumn{2}{c}{Long-context} \\
        \cmidrule(lr){3-9} \cmidrule(lr){10-11}
         & & \multicolumn{2}{c}{Spatial} & \multicolumn{2}{c}{Temporal} & \multicolumn{3}{c}{Partially Relevant} & \multicolumn{2}{c}{} \\
        \cmidrule(lr){3-4} \cmidrule(lr){5-6} \cmidrule(lr){7-9}
        & Method & CRB-S & VDC-O & CRB-T & CMRB & DREAM-E & LoVR-TH & PEV-K & LoVR-V & VDC-D \\
        & & R@1 & R@1 & R@1 & R@10 & R@1 & R@10 & R@1 & R@1 & R@1 \\
        \midrule

        \addlinespace[4pt]
        \multirow{5}{*}{\rotatebox[origin=c]{90}{\textit{MLLM-based}}}
        & GME-2B & 76.8 & 72.7 & 42.9 & 26.7 & 25.2 & 40.7 & \textbf{36.9} & 19.0 & 85.3 \\
        & UNITE-2B & 80.9 & 76.3 & 47.0 & 30.3 & 27.6 & 49.2 & \underline{36.4} & \textbf{82.0} & 83.1 \\
        & VLM2Vec-V2 & 79.8 & 76.9 & 30.2 & 27.5 & 23.3 & 47.8 & 24.0 & 63.0 & \underline{87.7} \\
        & BGE-VL & 86.5 & 76.6 & 44.8 & 31.2 & 25.2 & 47.9 & 30.4 & \underline{73.0} & 85.8 \\
        & GVE-3B & 83.9 & 76.5 & 47.4 & \textbf{36.2} & 27.7 & \underline{56.4} & 33.8 & 67.0 & \textbf{91.0} \\

        \midrule

        \multirow{6}{*}{\rotatebox[origin=c]{90}{\textit{CLIP-style}}}
        & CLIP4Clip & 49.6 & 62.3 & 30.4 & 27.6 & 19.8 & 34.8 & 17.9 & 32.0 & 55.3 \\
        & ViCLIP-L & 44.3 & 54.0 & 35.6 & 23.4 & 24.1 & 21.3 & 7.7 & 23.0 & 38.2 \\
        & VideoCLIP-XL & 84.0 & 72.9 & 48.7 & 27.3 & 26.7 & 43.7 & 22.7 & 38.0 & 81.7 \\
        & UMT-L & 44.1 & 57.2 & 34.0 & 32.6 & 21.4 & 23.3 & 4.0 & 26.0 & 47.0 \\
        & LanguageBind & 71.3 & 76.8 & 46.2 & 30.2 & 28.8 & 42.6 & 30.3 & 58.0 & 66.1 \\
        & InternVideo2-1B & 57.4 & 64.2 & 46.7 & \underline{36.0} & 24.3 & 30.6 & 2.7 & 26.0 & 47.9 \\

        \midrule
        
        \multirow{4}{*}{\rotatebox[origin=c]{90}{\textit{\textbf{Ours}}}}
        & \textbf{\baselinename-B} & 85.6 & 76.5 & 54.5 & 31.1 & 29.5 & 41.0 & 16.2 & 42.0 & 78.3 \\
        & \textbf{\methodname-B} & \underline{89.0} & 81.9 & 58.4 & 32.4 & \underline{34.8} & 55.3 & 10.8 & 60.0 & 87.1 \\
        & \textbf{\baselinename-L} & 87.8 & \underline{82.3} & \underline{60.8} & 32.4 & 34.1 & 48.9 & 20.8 & 49.0 & 76.9 \\
        & \textbf{\methodname-L} & \textbf{90.9} & \textbf{83.1} & \textbf{62.1} & 34.2 & \textbf{38.7} & \textbf{60.7} & 8.4 & 65.0 & 84.0 \\
        \bottomrule
    \end{tabular*}
\end{table*}

\paragraph{Fine-grained results}
The left part of \Cref{tab:retrieval_finegrained} shows the results for fine-grained video retrieval for spatial, temporal, and partially relevant queries. Training with multiple captions results in substantial gains on spatial and temporal retrieval. Relative to ViCLIP-L, \mbox{\baselinename-L} raises R@1 from 44.3\% to 87.8\% on CRB-S and from 35.6\% to 60.8\% on CRB-T. It already outperforms VideoCLIP-XL on all four spatial and temporal settings, with its largest R@1 gains on CRB-T (+12.1 points) and VDC-O (+9.4 points). We observe a similar trend on partially relevant retrieval with  DREAM-E and LoVR-TH. Finally, fine-grained alignment \mbox{\methodname} further improves on all fine-grained settings compared to the \mbox{\baselinename}, except for PEV-K (16.2\% to 10.8\% for ViT-B and 20.8\% to 8.4\% for ViT-L). This dataset is particularly difficult for CLIP-style video-text models because the text query is the concatenation of single-word concepts. A detailed analysis is provided in the Appendix. 

Among all evaluated models, \mbox{\methodname-L} achieves the best result on five of the seven fine-grained settings. On CRB-S, it raises the best prior R@1 from 86.5\% (BGE-VL) to 90.9\%. Its largest R@1 gains over prior models are on CRB-T (+13.4 points over VideoCLIP-XL) and DREAM-E (+9.9 points over LanguageBind). On LoVR-TH, it reaches 60.7\% R@10, 4.3 points above GVE-3B. GVE-3B leads CMRB, while GME-2B leads PEV-K. These results demonstrate the benefit of training with multiple captions and fine-grained alignment for queries describing specific spatial details, temporal changes, or events in only part of a video, compared to even 1B video encoder trained in multiple stages (InternVideo2-1B) and larger MLLM-based models trained for general multimodal retrieval (GME-2B, BGE-VL, UNITE-2B, and VLM2Vec-V2) or specifically for video retrieval (GVE-3B).

\paragraph{Long-context results}
The right part of \Cref{tab:retrieval_finegrained} shows that our \methodname also improves the performance on both long-context benchmarks.  R@1 increases from 49.0\% to 65.0\% on LoVR-V and from 76.9\% to 84.0\% on VDC-D, gains of 16.0 and 7.1 points, respectively. \mbox{\methodname-L} surpasses the compared CLIP-style methods on both benchmarks, improving the previous best scores from 58.0\% (LanguageBind) to 65.0\% on LoVR-V, and from 81.7\% (VideoCLIP-XL) to 84.0\% on VDC-D. Among the MLLM-based models, \mbox{\methodname-L} remains competitive, reducing the gap to GVE-3B to 2.0 points on LoVR-V and 7.0 points on VDC-D. Surprisingly, \mbox{\methodname-B} outperforms \mbox{\methodname-L} by 3.1 points on VDC-D. These gains show that the proposed method also benefits retrieval with long-context descriptions, although GVE-3B remains stronger in this setting.

\subsection{Ablations}
\label{sec:ablations}
To evaluate the different components of \methodname, we conduct a series of experiments with our ViT-B variant on a 1M-video subset in this section. We evaluate R@1 for T2V zero-shot retrieval on a subset of benchmarks to assess both coarse-grained and fine-grained performance. We conduct the experiments of \Cref{tab:ablation_frametoken,tab:fine_grained_interaction,tab:ablation_subcaption} without using the register tokens. When not specified, we use the vision [\texttt{CLS}] representation to compute the similarity score.

\begin{table}[t]
    \centering
    \caption{\textbf{Ablation of the main components of our method.} Subcap. loss: subcaption alignment loss; Temp. attn.: temporal attention across frame-tokens; Inference: average the [\texttt{CLS}] and maximum frame-token similarities (rows above use the [\texttt{CLS}] only). Bold denotes the best result.}
    \footnotesize
    \setlength{\tabcolsep}{5pt}
    \resizebox{\columnwidth}{!}{%
    \begin{tabular}{@{}lcccc@{}}
        \toprule
        Method & MSR-VTT & CRB-T & VDC-O & DREAM-E \\
        \midrule
        Global loss (\baselinename) & 41.3 & 48.7 & 76.1 & 26.6 \\
        \quad + Subcap.\ loss  & 43.3 & 51.0 & 78.1 & 28.4 \\
        \quad + Temp.\ attn.   & 42.9 & 51.1 & 78.7 & 28.4 \\
        \quad + Inference      & 42.5 & 51.4 & \textbf{78.8} & 30.1 \\
        \rowcolor{gray!15}
        \quad + Registers (\methodname) & \textbf{43.9} & \textbf{51.8} & 78.4 & \textbf{30.5} \\
        \bottomrule
    \end{tabular}%
    }
    \label{tab:ablation_main}
\end{table}

\paragraph{From \baselinename to \methodname}
\Cref{tab:ablation_main} reports the improvements obtained by progressively adding the different components of our method starting from the baseline \baselinename. The largest improvements comes from adding the alignment between the frame-tokens and the subcaptions, improving by 2 points on MSR-VTT. We observe similar gains on fine-grained benchmarks CRB-T and DREAM-E with improvements of 2.3 and 1.8 points, respectively. Adding the temporal attention block further improves performance on CRB-T and VDC-O. At inference, using both the [\texttt{CLS}] and the frame-token with the highest score similarity consistently improves the results on all fine-grained benchmarks. This highlights the complementary information captured by frame-tokens for fine-grained retrieval. Interestingly, register tokens improve R@1 on the temporal retrieval benchmark CRB-T and the event-based retrieval benchmark DREAM-E, from 51.4\% to 51.8\% and from 30.1\% to 30.5\%, respectively. We hypothesize that register tokens absorb textual information that cannot be attributed to specific frames, reducing irrelevant supervision on frame-tokens and enabling them to learn more accurate fine-grained representations. Following paragraphs further investigate the contribution of each component.


\begin{table}[t]
    \centering
    \caption{\textbf{Ablation study of frame-token alignment and attention masking.} (a) --: no subcaption alignment (global loss only); \texttt{CLS}: alignment between the global video representation and subcaptions; \texttt{FrTok}: alignment between frame-tokens and subcaptions. Bold denotes the best result.}
    \label{tab:ablation_frametoken}

    \begin{subtable}[b]{0.49\linewidth}
        \centering
        \resizebox{\linewidth}{!}{
        \begin{tabular}{l|c|c|c}
            \toprule
            Alignment & MSR-VTT & CRB-T & VDC-O \\
            \midrule
            --
                & 41.3 & 48.7 & 76.1 \\
            \texttt{CLS}
                & \textbf{42.9} & 50.6 & 77.1 \\
            \rowcolor{green!15}
            \texttt{FrTok}
                & \textbf{42.9} & \textbf{51.1} & \textbf{78.7} \\
            \bottomrule
        \end{tabular}}
        \caption{Effect of frame-token alignment}
        \label{tab:ablation_frametoken_a}
    \end{subtable}
    \hfill
    \begin{subtable}[b]{0.49\linewidth}
        \centering
        \resizebox{\linewidth}{!}{
        \begin{tabular}{c|c|c|c}
            \toprule
            Attn Mask & MSR-VTT & CRB-T & VDC-O \\
            \midrule
            $\times$
                & 42.8 & 50.5 & 76.6 \\
            \rowcolor{green!15}
            \checkmark
                & \textbf{42.9} & \textbf{51.1} & \textbf{78.7} \\
            \bottomrule
        \end{tabular}}
        \caption{Effect of attention masking}
        \label{tab:ablation_frametoken_b}
    \end{subtable}
\end{table}

\paragraph{Frame-token Alignment}
\Cref{tab:ablation_frametoken} analyzes the impact of the frame-tokens introduced in our method. In particular, \Cref{tab:ablation_frametoken_a} compares our frame-token alignment with a first approach that directly aligns subcaptions with the global [\texttt{CLS}]. This naive approach already improves performance on all benchmarks. We observe that using frame-tokens further increases R@1 from 50.6\% to 51.1\% on CRB-T and from 77.1\% to 78.7\% on VDC-O, while preserving performance on MSR-VTT. These results suggest that the granularity of subcaptions is better suited to frame-tokens than the global [\texttt{CLS}] representation.

Table~\ref{tab:ablation_frametoken_b} shows the impact of our attention mask applied to frame-tokens. Adding the attention mask brings consistent improvements, increasing  R@1 by 0.1 point on MSR-VTT, 0.6 point on CRB-T, and 2.1 points on VDC-O. These gains show the importance of restricting frame-token attention to the corresponding video frames, particularly for fine-grained tasks. 

\begin{table}[t]
    \centering
    \caption{\textbf{Effect of fine-grained interaction.} \textit{F2S}: Frame-to-Subcaption; \textit{S2F}: Subcaption-to-Frame; \textit{F2S +S2F}: combines both \textit{F2S} and \textit{S2F}. Bold denotes the best result.
    }
    \label{tab:fine_grained_interaction}
    \resizebox{0.7\linewidth}{!}{
    \begin{tabular}{l|c|c|c}
        \toprule
        Fine-grained Interaction & MSR-VTT & CRB-T & VDC-O \\
        \midrule
        F2S
            & 42.2 & 48.5 & 77.5 \\
        F2S + S2F
            & 41.1 & 50.7 & 76.7 \\
        \rowcolor{green!15}
        S2F (ours)
            & \textbf{42.9} & \textbf{51.1} & \textbf{78.7} \\
        \bottomrule
    \end{tabular}
    }
\end{table}

\paragraph{Fine-grained Interaction}
Our fine-grained alignment matches each subcaption to its most similar frame-token. An intuitive alternative would be to reverse the alignment direction, matching each frame-token to its most similar subcaption. To investigate this alternative, Table~\ref{tab:fine_grained_interaction} compares three different alignment strategies for the subcaption loss $\mathcal{L}_{\mathrm{sub}}$: (1) \textit{F2S}, where each frame-token is aligned with its most similar subcaption; (2) \textit{S2F}, where each subcaption is aligned with its most similar frame-token (our approach); and (3) \textit{F2S + S2F}, which combines both alignment directions, using the \textit{S2F} and \textit{F2S} similarities for the text-to-video and video-to-text losses, respectively. We observe that S2F surpasses both alternatives on all benchmarks. Compared to F2S, S2F improves R@1 by 2.6 points on CRB-T and 1.2 points on VDC-O, suggesting that the reverse alignment is less effective. We observe similar gains compared to F2S + S2F, with R@1 increasing from 41.1\% to 42.9\% on MSR-VTT and from 76.7\% to 78.7\% on VDC-O. We hypothesize that a subcaption describes an event that occurs in at least one frame, making it suitable to match it to its most similar frame-token. However, the reverse process is not always true, as some frame-tokens may represent visual content not described by any subcaption, leading to noisy supervision.

\begin{table}[t]
    \centering
\caption{\textbf{Ablation study of subcaption generation.} When not specified, we use $K=12$ subcaptions and sample $N=3$ maximum number of sentences per subcaption. Bold denotes the best and underline second-best result.}
    \label{tab:ablation_subcaption}

    \small
    \setlength{\tabcolsep}{3pt}

    \begin{subtable}[t]{0.60\columnwidth}
        \centering
        \resizebox{\linewidth}{!}{%
            \begin{tabular}{l|c|c|c}
                \toprule
                Caption source & MSR-VTT & CRB-T & VDC-O \\
                \midrule
                Single  & 42.8 & 49.7 & 76.3 \\
                \rowcolor{green!15}
                Multiple & \textbf{42.9} & \textbf{51.1} & \textbf{78.7} \\
                \bottomrule
            \end{tabular}%
        }
        \caption{Effect of combining multiple detailed captions for a single video.}
        \label{tab:ablation_subcaption_a}
    \end{subtable}

    \par\vspace{5pt}

    \begin{subtable}[t]{0.49\columnwidth}
        \centering
        \resizebox{\linewidth}{!}{%
            \begin{tabular}{c|c|c|c}
                \toprule
                \makebox[1.5cm][c]{\# Subcap.}
                    & MSR-VTT & CRB-T & VDC-O \\
                \midrule
                2  & 41.2 & 50.0 & 76.3 \\
                8  & 41.3 & 50.2 & \textbf{78.9} \\
                \rowcolor{green!15}
                12 & \textbf{42.9} & \underline{51.1} & \underline{78.7} \\
                16 & \underline{42.7} & \textbf{51.2} & 77.2 \\
                \bottomrule
            \end{tabular}%
        }
        \caption{Number of subcaptions $K$.}
        \label{tab:ablation_subcaption_b}
    \end{subtable}%
    \hfill
    \begin{subtable}[t]{0.49\columnwidth}
        \centering
        \resizebox{\linewidth}{!}{%
            \begin{tabular}{c|c|c|c}
                \toprule
                \makebox[1.5cm][c]{\# Sent.}
                    & MSR-VTT & CRB-T & VDC-O \\
                \midrule
                1    & 41.7 & \textbf{51.2} & 77.7 \\
                1--2 & 41.8 & 50.8 & \textbf{78.9} \\
                \rowcolor{green!15}
                1--3 & \underline{42.9} & \underline{51.1} & \underline{78.7} \\
                1--4 & \textbf{43.8} & 49.7 & 77.2 \\
                \bottomrule
            \end{tabular}%
        }
        \caption{Max. number of sentences $N$.}
        \label{tab:ablation_subcaption_c}
    \end{subtable}
\end{table}

\paragraph{Effect of the Subcaptions}
Table~\ref{tab:ablation_subcaption} presents several setups for our subcaption sampling strategy. First, \Cref{tab:ablation_subcaption_a} shows the benefit of combining different captions of the same video to form subcaptions. The gains are particularly noticeable on the fine-grained benchmarks CRB-T and VDC-O, where R@1 improves by 1.4 and 2.4 points, respectively. This approach allows us to generate more diverse subcaptions and further improve fine-grained textual supervision.

Second, \Cref{tab:ablation_subcaption_b} examines the number of sampled subcaptions $K$ during pretraining. We train our method with $K \in \{2, 8, 12, 16\}$. We observe that the $K=12$ setting provides the best overall trade-off across the three benchmarks, achieving the highest R@1 on MSR-VTT (42.9\%) and the second-highest on CRB-T (51.1\%) and VDC-O (78.7\%). For CRB-T and VDC-O, it falls only 0.1 and 0.2 point short of the best-performing settings, respectively.

Finally, \Cref{tab:ablation_subcaption_c} evaluates the influence of the maximum number of sentences for $N \in \{1, 2, 3, 4\}$. Compared to using individual sentences ($N=1$), our approach ($N=3$) improves R@1 by 1.2 points on MSR-VTT and 1.0 point on VDC-O. On CRB-T, both settings achieve similar performance, with a slight advantage for $N=1$ (51.2\% vs. 51.1\%). These results suggest that allowing subcaptions to contain multiple sentences provides a better supervision.
Overall, our method uses $K=12$ and $N=3$.


\begin{table}[t]
    \centering
    \vspace{-0.5cm}
    \caption{\textbf{Ablation study of inference scoring strategies.} We compare using the global [\texttt{CLS}] similarity or the best-maximum frame-token \texttt{FrTok} similarity alone, as well as combining both similarities using their maximum or mean. All scores are obtained with the same model (last row of Tab.~\ref{tab:ablation_main}). Bold denotes the best and underline second-best result.}
    \label{tab:inference}
    \resizebox{0.7\linewidth}{!}{
    \begin{tabular}{l|c|c|c}
        \toprule
        Inference & MSR-VTT & CRB-T & VDC-O \\
        \midrule
        \texttt{CLS} only
            & \underline{43.8} & 50.6 & 77.7 \\
        \texttt{FrTok} only
            & 37.0 & 48.8 & 77.9 \\
        max(\texttt{CLS}, \texttt{FrTok})
            & 43.6 & \underline{50.8} & \textbf{78.5} \\
        \rowcolor{green!15}
        mean(\texttt{CLS}, \texttt{FrTok})
            & \textbf{43.9} & \textbf{51.8} & \underline{78.4} \\
        \bottomrule
    \end{tabular}
    }
\end{table}

\paragraph{Retrieval Similarity Score}
At inference, we rank videos for a text query using the average of the global and fine-grained cosine similarities:
\begin{equation}
s(V,T)
=
\frac{1}{2}\left[
\mathrm{sim}(v_{\mathrm{cls}},t_{\mathrm{eos}})
+
\operatorname*{MaxSim}_{\ell}(f_\ell,t_{\mathrm{eos}})
\right].
\end{equation}
This assigns equal weight to global video semantics and the frame-token representation that best matches the query.
\Cref{tab:inference} compares alternative retrieval scoring strategies.
Rows 1 and 2 use only the global similarity or the maximum frame-token similarity, respectively.
Rows 3 and 4 combine these two scores by taking their maximum or their average, respectively. We see that replacing the global [\texttt{CLS}] token similarity with the maximum frame-token one degrades performance on almost all benchmarks, with a significant drop of 6.8 points in R@1 on MSR-VTT. This drop is expected, as MSR-VTT evaluates global representations, whereas frame-tokens capture only local visual information. In most cases, combining both similarities outperforms their individual counterpart, indicating that the global [\texttt{CLS}] token and the best-matching frame-token capture complementary information. Specifically, row 4 shows the largest gains against row 1 on fine-grained benchmarks, improving R@1 from 50.6\% to 51.8\% on CRB-T and from 77.7\% to 78.4\% on VDC-O. Overall, these results demonstrate the benefit of using the maximum frame-token similarity during inference to leverage fine-grained representations.

\begin{figure}[t]
    \centering
    \vspace{-0.5cm}
    \includegraphics[width=\linewidth]{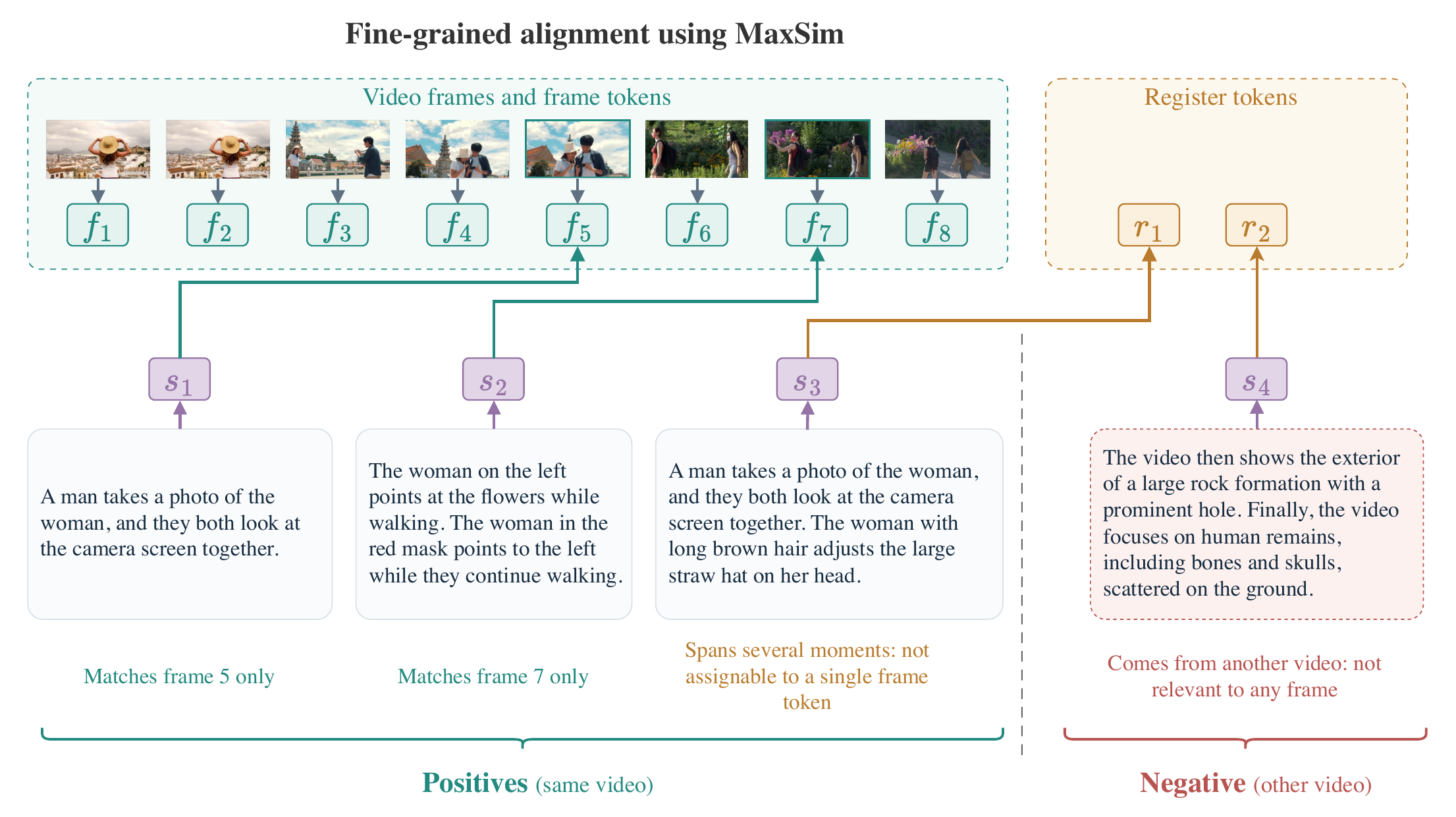}
    \vspace{-0.5cm}
    \caption{\textbf{Qualitative example of our MaxSim-based fine-grained alignment.}
    Each subcaption is matched to its most similar token among the frame-tokens and register tokens. Subcaptions describing frame-specific content are matched to frame-tokens, while those containing global information or content unrelated to the video are matched to register tokens.}
    \label{fig:example}
\end{figure}

\subsection{Qualitative Example}
\Cref{fig:example} illustrates the alignment between frame-tokens, register tokens, and a subset of subcaptions. Subcaptions $s_1$ and $s_2$ are correctly aligned with their corresponding frame-tokens, $f_5$ and $f_7$, respectively. In contrast, subcaption $s_3$ is aligned with a register token, as it combines two sentences describing content from frames 1 and 3. Individually, these sentences would be expected to align with frame-tokens $f_3$ and $f_1$, respectively. However, their concatenation spans multiple frames and cannot be unambiguously attributed to a single frame-token. This example highlights the role of register tokens in capturing visual information that extends across multiple frames. Finally, subcaption $s_4$, drawn from another video sample, is also correctly matched to a register token. Importantly, register tokens are not assigned predefined alignments for positive or negative pairs.

\section{Conclusion}
\label{sec:conclusion}
We present \methodname, a video-text pretraining method that jointly learns global and fine-grained video representations. We first introduce \baselinename, a strong baseline that leverages multiple summary and detailed captions through global contrastive alignment. Building on this baseline, \methodname includes learnable frame-tokens and register tokens to capture complementary visual information, together with a subcaption sampling strategy that provides textual supervision at different levels of granularity. Our fine-grained contrastive objective aligns each subcaption with its most relevant visual token without requiring explicit temporal supervision. Extensive zero-shot text-to-video retrieval experiments demonstrate that \methodname achieves state-of-the-art performance on several coarse-grained and fine-grained benchmarks. Future work might explore additional training objectives beyond contrastive learning to further exploit detailed captions.

\section*{Acknowledgment}
Renaud Vandeghen is a Postdoctoral Researcher of the Fonds de la Recherche Scientifique – FNRS. The present research benefited from computational resources made available on Lucia, the Tier-1 supercomputer of the Walloon Region, infrastructure funded by the Walloon Region under the grant agreement n°1910247. We acknowledge LUMI-BE for awarding this project access to the LUMI supercomputer, owned by the EuroHPC Joint Undertaking, hosted by CSC (Finland) and the LUMI consortium through a LUMI-BE Regular Access call. LUMI-BE is joint effort from BELSPO (federal), SPW Économie, Emploi, Recherche (Wallonia), Department of Economy, Science \& Innovation (Flanders) and Innoviris (Brussels). The research reported in this publication was supported by funding from King Abdullah University of Science and Technology (KAUST) - Center of Excellence for Generative AI, under award number 5940. For computing time, this research used Ibex managed by the Supercomputing Core Laboratory at King Abdullah University of Science \& Technology (KAUST) in Thuwal, Saudi Arabia.

\clearpage
\setcounter{page}{1}
\maketitlesupplementary

\appendix
\label{sec:appendix}

\Cref{sec:evaluation} describes the datasets, retrieval protocol,
and model checkpoints used in our unified evaluation.
\Cref{sec:add-results} presents a broader comparison with published
zero-shot retrieval results, along with fine-tuned text-to-video
retrieval and zero-shot action classification results.
\Cref{sec:pev-k} discusses PEV-K performance.
\Cref{sec:longviclip} describes how we adapt ViCLIP into \baselinename.
\Cref{sec:add-ablations} provides an additional ablation study on frame tokens and register tokens.
\Cref{sec:training} provides the pretraining and fine-tuning hyperparameters. Finally, \Cref{sec:qual} shows qualitative examples of token-subcaption alignment, generated subcaptions, and fine-grained text-to-video retrieval.

\section{Unified Evaluation Protocol}
\label{sec:evaluation}
We present the controlled evaluation protocol used to obtain the zero-shot text-to-video retrieval results in~\Cref{tab:retrieval_standard,tab:retrieval_finegrained}. Our goal is to enable researchers to reproduce these results and compare new methods with prior work under the same conditions. The protocol has three components. First, we fix one evaluation split per benchmark and use its Hugging Face revision when available (\Cref{sec:datasets}). Second, we use a common retrieval procedure across all 15 tasks (\Cref{sec:protocol}). Third, we use the open-weight checkpoints used for prior methods and document the inference settings for every model (\Cref{sec:models}).
We plan to release this evaluation protocol publicly.

\subsection{Datasets}
\label{sec:datasets}
We use the 12 text-to-video tasks from the public UVRB release~\cite{Guo2025Towards-arxiv} (Alibaba-NLP/UVRB, revision \texttt{2b5ec76}). We additionally evaluate ActivityNet Captions, MSVD, and LSMDC using their retrieval annotations.

\paragraph{MSRVTT}
The MSR-VTT~\cite{Xu2016MSRVTT} test set contains 1,000 videos from the 1K-A split, each paired with one human-written caption describing the clip as a whole.

\paragraph{DiDeMo}
The DiDeMo~\cite{Hendricks2017Localizing} test set contains 1,004 videos and one paragraph query per video. Each paragraph concatenates sentences describing individual moments in the same video.

\paragraph{ANet}
The ActivityNet Captions~\cite{Caba2015ActivityNet} validation set used here contains 4,917 untrimmed videos from the first validation split. We concatenate each video's event descriptions in annotation order to form one paragraph query per video.

\paragraph{MSVD}
The MSVD~\cite{Chen2011Collecting} test set contains 670 videos and 27,763 English captions. Multiple annotators describe the same clip independently, and each caption serves as a separate query.

\paragraph{LSMDC}
The LSMDC~\cite{Rohrbach2015ADataset} public test set from the 2016 challenge contains 1,000 movie clips, each paired with one sentence describing characters, actions, or interactions within the scene.

\paragraph{CRB-G}
The CaReBench-General~\cite{Xu2026CaReBench} test set contains 1,000 videos, each paired with a detailed caption describing both the appearance of the scene and the actions that unfold.

\paragraph{CRB-S}
The CaReBench-Spatial~\cite{Xu2026CaReBench} test set contains the same 1,000 videos as CRB-G, each paired with a spatial caption describing visible entities, their appearance, and their relative positions.

\paragraph{VDC-O}
The VDC-Object~\cite{Chai2025AuroraCap} test set contains 1,027 videos, each paired with a description of its principal subject. The text identifies the subject through visible attributes such as appearance or clothing.

\paragraph{CRB-T}
The CaReBench-Temporal~\cite{Xu2026CaReBench} test set contains the same 1,000 videos as CRB-G, each paired with a temporal caption describing the order of actions or changes in the state of a subject or object.

\paragraph{CMRB}
The CameraBench~\cite{Lin2025Towards} test set used for retrieval contains 728 camera-motion descriptions and 1,071 candidate clips. Queries describe how the camera moves during a shot, for example by panning or moving toward a subject.

\paragraph{DREAM-E}
The DREAM-1K-Event~\cite{Wang2024Tarsier-arxiv} test set contains 6,251 event descriptions and 1,000 candidate videos. Each query describes one event within a clip, so distinct queries can refer to different moments in the same video.

\paragraph{LoVR-TH}
The LoVR-Theme2Clip~\cite{Liang2026LoVR} test set contains 8,854 clips, each paired with a description of its overall theme, mood, or narrative style.

\paragraph{PEV-K}
The PE-Video-Keyword~\cite{Bolya2025Perception} test set contains 14,427 keyword queries and 15,000 candidate videos. Each query is a comma-separated list of tags describing objects, activities, settings, or visual attributes.

\paragraph{LoVR-V}
The LoVR-Text2Video~\cite{Liang2026LoVR} test set contains 100 long descriptions and 467 full-length candidate videos, including 367 distractors. Each query describes content across multiple scenes in its target video.

\paragraph{VDC-D}
The VDC-Detail~\cite{Chai2025AuroraCap} test set contains 1,027 detailed captions paired with the same 1,027 videos used in VDC-O. Each caption describes the subjects, setting, camera movement, and sequence of events throughout the clip.

\subsection{Evaluation protocol}
\label{sec:protocol}
We use MTEB~\cite{Muennighoff2023MTEB} (version 2.26.33) to sample frames uniformly across each video. We encode text and video with CUDA BF16 autocast while retaining each model's native weight precision. All reported results were processed with an NVIDIA A100 40GB GPU. 

\subsection{Models}
\label{sec:models}
We evaluate the released checkpoints of prior methods and our own training checkpoints. \Cref{tab:supp_retrieval_models} identifies the configurations used for the reported scores.

\begin{table}[t]
\centering
\small
\setlength{\tabcolsep}{4pt}
\caption{\textbf{Configurations for text-to-video retrieval.} Checkpoint names identify the evaluated weights on Hugging Face using their organization/model identifiers.}
\label{tab:supp_retrieval_models}
\begin{tabular}{@{}ll@{}}
\toprule
Method & Checkpoint or backbone \\
\midrule
CLIP4Clip~\cite{Luo2022CLIP4Clip} & Searchium-ai/clip4clip-webvid150k \\
ViCLIP-L~\cite{Wang2024InternVid} & OpenGVLab/ViCLIP-L-14-hf \\
VideoCLIP-XL~\cite{Wang2024VideoCLIPXL} & alibaba-pai/VideoCLIP-XL \\
UMT-L~\cite{Li2023Unmasked} & OpenGVLab/UMT-L16-17m \\
LanguageBind~\cite{Zhu2024LanguageBind} & \shortstack[t]{LanguageBind/\\LanguageBind\_Video\_Huge\_V1.5\_FT} \\
InternVideo2-1B~\cite{Wang2024InternVideo2} & \shortstack[t]{OpenGVLab/\\InternVideo2-Stage2\_1B-224p-f4} \\
GVE-3B~\cite{Guo2025Towards-arxiv} & Alibaba-NLP/GVE-3B \\
GME-2B~\cite{Zhang2025Bridging} & Alibaba-NLP/gme-Qwen2-VL-2B-Instruct \\
UNITE-2B~\cite{Kong2025Modality-arxiv} & friedrichor/Unite-Base-Qwen2-VL-2B \\
VLM2Vec-V2~\cite{Meng2026VLM2VecV2} & VLM2Vec/VLM2Vec-V2.0 \\
BGE-VL~\cite{Zhou2025MegaPairs} & BAAI/BGE-VL-v1.5-mmeb \\
\baselinename-B/-L & ViT-B/16 / ViT-L/14 \\
\methodname-B/-L & ViT-B/16 / ViT-L/14 \\
\bottomrule
\end{tabular}
\end{table}

\section{Additional Results}
\label{sec:add-results}

\begin{table*}[t]
    \centering
    \caption{\textbf{Broader comparison on zero-shot text-to-video retrieval.}
    We evaluate our methods against state-of-the-art methods on coarse-grained zero-shot retrieval benchmarks.
    Img15M denotes CC12M + SBU + COCO + VG. CC denotes Conceptual Captions, and VG denotes Visual Genome. Bold denotes the best and underline second-best result.}
    \label{tab:retrieval_sota}
    \scriptsize
    \setlength{\tabcolsep}{1.5pt}
    \renewcommand{\arraystretch}{1.0}
    \begin{tabularx}{\textwidth}{@{}cl>{\centering\arraybackslash}X
    *{5}{>{\centering\arraybackslash}p{0.075\textwidth}}@{}}

        \toprule
        & Method & {Pretraining data}
        & MSRVTT & DiDeMo & ANet & MSVD & LSMDC \\
        & & & R@1 & R@1 & R@1 & R@1 & R@1 \\
        \midrule

        \multicolumn{8}{@{}c@{}}{\textbf{Cross-encoder}} \\

        \rowcolor{crossencodergray}
        & VIOLET~\cite{Fu2021VIOLET-arxiv} & {YT-Temporal + WebVid-2M + CC3M}
        & 25.9 & 23.5 & -- & -- & -- \\

        \rowcolor{crossencodergray}
        & Singularity~\cite{Lei2023Revealing} & WebVid-2M + Img15M
        & 34.0 & 37.1 & 30.6 & -- & -- \\
        \rowcolor{crossencodergray}
        & UMT-L~\cite{Li2023Unmasked} & WebVid-2M + Img15M
        & 42.6 & 46.4 & 42.8 & 49.9 & 25.2 \\
        \rowcolor{crossencodergray}
        & UMT-L (Panda)~\cite{Chen2024Panda70M} & Panda-5M
        & 37.2 & 34.2 & -- & 37.2 & -- \\
        \rowcolor{crossencodergray}
        & OmniVL~\cite{Wang2022OmniVL} & WebVid-2.5M + Img14M
        & 34.6 & 33.3 & -- & -- & -- \\
        \rowcolor{crossencodergray}
        & InternVideo2$_{s2}$-1B~\cite{Wang2024InternVideo2} & IV-25.5M
        & 51.9 & 57.0 & 60.4 & 58.1 & 32.0 \\
        \rowcolor{crossencodergray}
        & MVC-L~\cite{Thoker2026Advancing-arxiv} & InternVid-10M-FLT
        & 41.1 & 62.0 & 66.9 & 49.0 & 24.1 \\

        \midrule

        \multicolumn{8}{@{}c@{}}{\textbf{Dual-encoder}} \\

        & Frozen~\cite{Bain2021Frozen} & WebVid-2M + Img3M
        & 23.2 & 21.1 & -- & -- & -- \\
        & CLIP~\cite{Radford2021Learning} & CLIP-400M
        & 30.4 & 12.7 & 9.1 & 40.5 & 13.9 \\
        & InternVideo~\cite{Wang2022InternVideo-arxiv} & WebVid-10M + HowTo100M + LAION-400M
        & 40.7 & 31.5 & 30.7 & 43.4 & 17.6 \\
        & VideoCoCa-L~\cite{Yan2022VideoCoCa-arxiv} & CoCa-4B + VideoCC3M
        & 33.9 & -- & 31.5 & -- & -- \\
        & CLIP4Clip~\cite{Luo2022CLIP4Clip} & CLIP-400M
        & 32.0 & -- & -- & 38.5 & 15.1 \\
        & ViCLIP-L~\cite{Wang2024InternVid} & CLIP-400M + InternVid-10M-FLT
        & 42.4 & 18.4 & 15.1 & 49.1 & 20.1 \\
        & VideoCLIP-XL~\cite{Wang2024VideoCLIPXL} & CLIP-400M + InternVid-10M-FLT + VILD2M
        & \textbf{50.1} & 47.7 & 46.4 & 51.9 & 22.8 \\
        & LanguageBind~\cite{Zhu2024LanguageBind} & VIDAL-10M
        & 42.6 & 37.8 & 35.1 & 52.2 & -- \\

        & \textbf{\baselinename-L} (ours)
        & CLIP-400M + InternVid-10M-FLT
        & 46.9 & \underline{52.5} & \textbf{55.2} & \underline{57.2} & \underline{24.3} \\
        & \textbf{\methodname-L} (ours)
        & CLIP-400M + InternVid-10M-FLT
        & \underline{48.8} & \textbf{56.3} & \underline{55.1} & \textbf{58.1} & \textbf{25.7} \\

        \bottomrule
    \end{tabularx}
\end{table*}
\label{sec:sota}
\subsection{Broader Comparison on Zero-shot Text-Video Retrieval}
\label{sec:retrieval_published}

Beyond our unified evaluation in \Cref{sec:evaluation}, which uses a common protocol and evaluation settings across models, \Cref{tab:retrieval_sota} broadens the comparison to include published results from both dual encoders and large cross-encoders, including UMT-L~\cite{Li2023Unmasked}, InternVideo2-1B~\cite{Wang2024InternVideo2}, MVC~\cite{Thoker2026Advancing-arxiv}. We report R@1 for prior methods as given in their original papers, under their respective evaluation settings.

Among the dual encoders, which follow the same retrieval setup as our models, either \mbox{\baselinename-L} or \mbox{\methodname-L} achieves the highest R@1 on all benchmarks except MSR-VTT. \mbox{\methodname-L} leads on DiDeMo (56.3\%), LSMDC (25.7\%), and MSVD (58.1\%), while \mbox{\baselinename-L} leads on ActivityNet Captions (55.2\%). On MSR-VTT, VideoCLIP-XL achieves the highest R@1 (50.1\%), followed by \mbox{\methodname-L} (48.8\%). These results show that our models achieve strong zero-shot retrieval performance while retaining the ability to precompute and reuse video embeddings across queries.

Cross-encoders use additional Transformer layers to model interactions between video and text features. These interactions can improve retrieval accuracy, but they come at a substantial inference cost. Even if the video features are precomputed, the cross-modal layers must process each candidate video again for every new text query. This makes exhaustive search expensive as the collection grows. To that end, on MSR-VTT, our method  achieves a retrieval speedup of approximately $4\times$ over UMT-L and MVC~\cite{Thoker2026Advancing-arxiv} and around  $8\times$ over InternVideo2-1B. With a dual-encoder architecture, \mbox{\methodname-L} exceeds the SOTA cross-encoders UMT-L and MVC on five and three retrieval datasets respectively. It also matches InternVideo2-1B on MSVD (58.1\%). This demonstrates that \mbox{\methodname-L} achieves strong zero-shot retrieval accuracy while retaining efficient inference through precomputed, reusable video embeddings.



\begin{table}[t]
    \centering
    \caption{\textbf{Fine-tuned text-to-video retrieval.}
    We evaluate our methods against state-of-the-art methods on fine-tuned retrieval benchmarks. For prior works, we report the results from~\cite{Wang2024VideoCLIPXL} using the same fine-tuning setup. Bold denotes the best and underline the second-best result.
    }
    \label{tab:finetuned_standard}
    \scriptsize
    \setlength{\tabcolsep}{1.5pt}
    \renewcommand{\arraystretch}{1.0}
    \begin{tabular*}{\columnwidth}{@{\extracolsep{\fill}}cl*{6}{c}@{}}
        \toprule
        & Method & MSRVTT & DiDeMo & ANet & MSVD & LSMDC \\
        & & R@1 & R@1 & R@1 & R@1 & R@1  \\
        \midrule

        \multirow{6}{*}{\rotatebox[origin=c]{90}{\textit{}}}
        & CLIP & 38.2 & 32.2 & 26.1 &  52.3 &  22.5 \\
        & CLIP4Clip & 45.6 & 43.0 & 40.3 & 45.2 & 24.3  \\
        & ViCLIP-L & 52.5 & 49.4 & 49.8 & 53.1 & 33.0  \\
        & VideoCLIP-XL & \underline{57.0} & 62.3 & 58.4 & 55.6 & 34.2\\

        \midrule

        \multirow{2}{*}{\rotatebox[origin=c]{90}{\textit{\textbf{Ours}}}}
        & \textbf{\baselinename-L} & \textbf{57.2} & \underline{67.4} & \underline{63.5} & \underline{60.8} & \underline{34.6} &  \\
        & \textbf{\methodname-L} & 55.4 & \textbf{69.2} & \textbf{64.3} & \textbf{61.5} & \textbf{35.9} \\
        \bottomrule
    \end{tabular*}
\end{table}

\subsection{Fine-tuned Text-to-Video Retrieval}
\label{sec:ft_t2v}

\Cref{tab:finetuned_standard} reports text-to-video R@1 after fine-tuning on each of the five benchmarks. We follow the fine-tuning setup of VideoCLIP-XL~\cite{Wang2024VideoCLIPXL} and use the prior results reported in ~\cite{Wang2024VideoCLIPXL}. \mbox{\baselinename-L} exceeds the best prior R@1 on all five benchmarks. \mbox{\methodname-L} further improves over \mbox{\baselinename-L} on DiDeMo, ActivityNet Captions, MSVD, and LSMDC, with gains of 1.8, 0.8, 0.7, and 1.3 points, respectively. The largest gains over VideoCLIP-XL are on DiDeMo (62.3\% to 69.2\%) and ActivityNet Captions (58.4\% to 64.3\%). On MSR-VTT, \mbox{\baselinename-L} achieves the highest R@1 at 57.2\%, while \mbox{\methodname-L} reaches 55.4\%. 
Similar to the zero-shot results, \mbox{\methodname-L} improves over prior best methods on four of the five benchmarks after full fine-tuning, with gains of up to 6.9 points in R@1.
\begin{table}[t]
    \centering
    \caption{\textbf{Zero-shot action classification.}
    We evaluate our methods against CLIP-based video-text models on zero-shot video classification benchmarks. We report top-1 accuracy computed using the [\texttt{CLS}] token.  All models are evaluated with the same protocol. Bold denotes the best and underline second-best result.
    }
    \label{tab:zeroshot_classification}
    \scriptsize
    \setlength{\tabcolsep}{1.5pt}
    \renewcommand{\arraystretch}{1.0}
    \begin{tabular*}{\columnwidth}{@{\extracolsep{\fill}}cl*{3}{c}@{}}
        \toprule
        & Method & K400 & K600 & K700 \\
        & & Acc. & Acc. & Acc. \\
        \midrule

        \multirow{3}{*}{\rotatebox[origin=c]{90}{\textit{ }}}
        & CLIP4Clip & 49.6 & 47.0 & 38.0 \\
        & ViCLIP-L & 59.9 & 57.9 & 49.9 \\
        & VideoCLIP-XL & 63.4 & 61.2 & 53.5 \\

        \midrule

        \multirow{2}{*}{\rotatebox[origin=c]{90}{\textit{\textbf{Ours}}}}
        & \textbf{\baselinename-L} & \underline{65.2} & \underline{62.6} & \underline{55.2} \\
        & \textbf{\methodname-L} (\texttt{CLS} only) & \textbf{65.3} & \textbf{63.6} & \textbf{56.0} \\

        \bottomrule
    \end{tabular*}
\end{table}
\subsection{Zero-Shot Action Classification}
\label{sec:zeroshot_classification}

\Cref{tab:zeroshot_classification} reports zero-shot top-1 accuracy on Kinetics-400~\cite{Carreira2017QuoVadis}, Kinetics-600~\cite{Carreira2018AShort-arxiv}, and Kinetics-700~\cite{Carreira2019AShort-arxiv} under a common evaluation protocol using MTEB~\cite{Muennighoff2023MTEB}. In particular, the task is formulated as video-to-text retrieval, where each video is a query, and the text class labels are the candidates. \mbox{\baselinename-L} outperforms all prior models in the table on all three benchmarks. \mbox{\methodname-L} further improves over \mbox{\baselinename-L} by 0.1, 1.0, and 0.8 points, respectively. These results show that fine-grained alignment during pretraining also benefits the global video representation for zero-shot action recognition.

\section{Discussion on PEV-K}
\label{sec:pev-k}

In~\Cref{tab:retrieval_finegrained}, most CLIP-style models, including our \baselinename and \methodname, achieve poor performance on the PEV-K benchmark. As described in \Cref{sec:datasets}, the 14,427 PEV-K queries consist of comma-separated keywords rather than natural descriptions. On average, each query contains $\sim$ 40 items out of the 46,311 ones. Representative examples include:

{\footnotesize
\begin{itemize}
    \setlength{\itemsep}{2pt}
    \setlength{\topsep}{3pt}
    \item \texttt{"table,\allowbreak
    bar,\allowbreak hands,\allowbreak Hand,\allowbreak
    casino,\allowbreak poker,\allowbreak barman,\allowbreak
    poker chips,\allowbreak poker chip"}

    \item \texttt{"art,\allowbreak
    people,\allowbreak turkey,\allowbreak workers,\allowbreak
    craft,\allowbreak processing,\allowbreak labor,\allowbreak
    psoriasis"}

    \item \texttt{"liquid,\allowbreak
    oil,\allowbreak black,\allowbreak replace,\allowbreak
    car,\allowbreak engine,\allowbreak repair,\allowbreak
    fluid,\allowbreak bolt,\allowbreak replacement,\allowbreak
    Carter"}
\end{itemize}
}
We hypothesize that this unusual query format introduces a significant distribution mismatch with the captions typically used during video-text pretraining.

\section{From ViCLIP to \baselinename}
\label{sec:longviclip}
\begin{table}[t]
    \centering
    \caption{\textbf{Adapting ViCLIP into \baselinename.}
    Each step adds the indicated change to the preceding configuration. The final \baselinename{} configuration is shaded in gray. Bold denotes the best result.
    }
    \label{tab:ablation_baseline}

    \resizebox{\linewidth}{!}{%
    \begin{tabular}{@{}lccc@{}}
        \toprule
        Adaptation step & MSR-VTT & CRB-T & VDC-O \\
        \midrule
        ViCLIP (1 short caption, 32 tokens)
        & 32.3 & 25.6 & 43.0 \\

        \quad + Extend text context (32 $\to$ 134 tokens)
        & 31.4 & 30.4 & 54.3 \\

        \quad + Add one long caption per video
        & 33.8 & 34.3 & 62.9 \\

        \quad + Reduce random masking (90\% $\to$ 50\%)
        & 36.8 & 37.1 & 66.2 \\

        \quad + Unfreeze text encoder (LR $2\times10^{-5}$)
        & 35.9 & 41.5 & 69.3 \\

        \midrule
        \rowcolor{gray!15}
        \quad + Use multiple short and long captions (\baselinename)
        & \textbf{41.4} & \textbf{46.4} & \textbf{75.2} \\
        \bottomrule
    \end{tabular}%
    }
\end{table}

\Cref{tab:ablation_baseline} shows the progression from the ViCLIP configuration to \baselinename. We start with the ViCLIP configuration~\cite{Wang2024InternVid}, which uses a single short caption per video, a frozen text encoder with a 32-token context, and a random masking ratio of 90\%. Each step builds on the previous configuration. We report zero-shot text-to-video R@1.

We first extend the text context from 32 to 134 tokens by linearly interpolating the positional embeddings, following \cite{Zhang2024LongCLIP,Thoker2026Advancing-arxiv}. This increases R@1 from 25.6\% to 30.4\% on CRB-T and from 43.0\% to 54.3\% on VDC-O. On MSR-VTT, R@1 decreases by 0.9 points. We then add one long caption per video~\cite{Thoker2026Advancing-arxiv} and align the global \texttt{[CLS]} representation with both the short and the long caption. This improves R@1 on all three benchmarks, with the largest gain on VDC-O (54.3\% to 62.9\%).

Next, we reduce random masking from 90\% to 50\%, exposing the video encoder to more patches during training. R@1 improves on all three benchmarks. We then unfreeze the text encoder and train it with a learning rate of $2\times10^{-5}$ to adapt it to the longer captions. R@1 increases by 4.4 points on CRB-T and 3.1 points on
VDC-O, and decreases by 0.9 points on MSR-VTT.

Finally, we train with multiple short and long captions per video to obtain \baselinename. This step improves R@1 on all three benchmarks, yielding 41.4\% on MSR-VTT, 46.4\% on CRB-T, and 75.2\% on VDC-O. Relative to the starting ViCLIP configuration, these scores represent gains of 9.1, 20.8, and 32.2 points, respectively. To build \methodname, we also increased the context length to 248 and reduced the masking to 10\%, which did not significantly improve performance.

\section{Additional Ablations}
\label{sec:add-ablations}
In this section, we present a series of additional ablations using the exact same setup as the previous main ablations. We  report zero-shot text-to-video (T2V) R@1. 

\subsection{Frame-token and Register token}
\label{sec:ablation-frametoken}
\begin{table}[t]
    \centering
    \caption{\textbf{Ablations on the video encoder.} (a) Number of frame tokens. (b) Number of registers. (c) Learnable frame tokens vs.\ frame tokens obtained by mean pooling. (d) Number of inter-frame attention layers. Bold denotes the best and underline second-best result}
    \label{tab:ablation_encoder}

    \small
    \setlength{\tabcolsep}{3pt}

    \begin{subtable}[t]{0.49\columnwidth}
        \centering
        \resizebox{\linewidth}{!}{%
            \begin{tabular}{c|c|c|c}
                \toprule
                \makebox[1.5cm][c]{\# Frame tok.}
                    & MSR-VTT & CRB-T & VDC-O \\
                \midrule
                \rowcolor{green!15}
                1 & \textbf{42.9} & \textbf{51.1} & \textbf{78.7} \\
                2 & 42.8 & 50.0 & 78.1 \\
                3 & 41.8 & 49.9 & 78.3 \\
                \bottomrule
            \end{tabular}%
        }
        \caption{Number of frame tokens.}
        \label{tab:ablation_encoder_a}
    \end{subtable}%
    \hfill
    \begin{subtable}[t]{0.49\columnwidth}
        \centering
        \resizebox{\linewidth}{!}{%
            \begin{tabular}{c|c|c|c}
                \toprule
                \makebox[1.5cm][c]{\# Reg.}
                    & MSR-VTT & CRB-T & VDC-O \\
                \midrule
                1 & \underline{43.3} & \textbf{51.2} & 76.8 \\
                \rowcolor{green!15}
                2 & \textbf{43.8} & \underline{50.6} & \underline{77.7} \\
                3 & 42.3 & \underline{50.6} & \textbf{78.0} \\
                \bottomrule
            \end{tabular}%
        }
        \caption{Number of registers.}
        \label{tab:ablation_encoder_b}
    \end{subtable}

    \par\vspace{5pt}

    \begin{subtable}[t]{0.49\columnwidth}
        \centering
        \resizebox{\linewidth}{!}{%
            \begin{tabular}{c|c|c|c}
                \toprule
                \makebox[1.5cm][c]{Frame tok.}
                    & MSR-VTT & CRB-T & VDC-O \\
                \midrule
                \rowcolor{green!15}
                Learnable & \textbf{42.9} & \textbf{51.1} & \textbf{78.7} \\
                Mean pool & 41.4 & 49.1 & 77.1 \\
                \bottomrule
            \end{tabular}%
        }
        \caption{Learnable vs.\ mean-pooled frame tokens.}
        \label{tab:ablation_encoder_c}
    \end{subtable}%
    \hfill
    \begin{subtable}[t]{0.49\columnwidth}
        \centering
        \resizebox{\linewidth}{!}{%
            \begin{tabular}{c|c|c|c}
                \toprule
                \makebox[1.5cm][c]{\# Attn. layers}
                    & MSR-VTT & CRB-T & VDC-O \\
                \midrule
                \rowcolor{green!15}
                2 & \textbf{42.9} & \textbf{51.1} & \textbf{78.7} \\
                4 & 42.0 & \textbf{51.1} & 77.6 \\
                \bottomrule
            \end{tabular}%
        }
        \caption{Number of inter-frame attention layers.}
        \label{tab:ablation_encoder_d}
    \end{subtable}
\end{table}

Table~\ref{tab:ablation_encoder} reports several ablations for our vision encoder. First, Table~\ref{tab:ablation_encoder_a} studies the number of frame-tokens per-frame in our vision encoder. Compared to our approach (one frame-token per frame), using two or three frame-tokens generally reduces performance. With two frame-tokens, T2V retrieval R@1 drops by 0.1 point on MSR-VTT, by 1.1 points on CRB-T and 0.6 points on VDC-O. With three frame-tokens, the drops are 1.1, 1.2 and 0.4 points, respectively. We therefore adopt one frame-token per frame in our setting.

Table~\ref{tab:ablation_encoder_b} examines the number of register tokens during pretraining. We see that using two register tokens is the best overall trade-off across the three benchmarks. It achieves the highest R@1 on MSR-VTT (43.8\% T2V) and CRB-T (55.5\% T2V). It ranks second in T2V, with a score just 0.3 points below the best result. We finally adopt two register tokens in our setting. 

Next, we investigate whether simply averaging the patches of each frame to form frame-tokens would lead to similar results. To investigate this, Table~\ref{tab:ablation_encoder_c} compares learnable frame-tokens with mean-pooled frame-tokens. It shows that learnable frame-tokens outperform mean pooling by 1.6 points on VDC-O and by 2.0 points on CRB-T. These results suggest that learnable frame-tokens may capture more fine-grained spatial information from each frame, whereas mean pooling may dilute such fine-grained information. 

Finally, in Table~\ref{tab:ablation_encoder_d}, we compare temporal attention blocks with two and four layers. The two-layer block generally achieves better results across the three benchmarks. We choose to use a two-layer temporal attention block in our setting.

\begin{table*}[t]
    \centering
    \caption{\textbf{Training hyperparameters.}
    We report the hyperparameters used to pretrain the base and large variants of \baselinename and \methodname, as well as those used to fine-tune both methods on video-text retrieval benchmarks.
    }
    \label{tab:training-hyperparameters}
    \small
    \setlength{\tabcolsep}{4pt}

    \newcommand{\sharedcell}[1]{%
        \multicolumn{4}{c}{\makebox[0pt][c]{#1}}%
    }

    \begin{subtable}{\textwidth}
        \centering
        \caption{Pretraining hyperparameters.}
        \label{tab:hyperparameters}
        \begin{tabular}{@{}lcccc@{}}
            \toprule
             & \multicolumn{2}{c}{\baselinename}
             & \multicolumn{2}{c}{\methodname} \\
            \cmidrule(lr){2-3}\cmidrule(lr){4-5}
            Hyperparameter & ViT-B/16 & ViT-L/14 & ViT-B/16 & ViT-L/14 \\
            \midrule
            Vision and text initialization
                & \sharedcell{CLIP, matched to backbone size} \\
            Video input
                & \sharedcell{\topstack{8 frames, one randomly sampled per temporal segment,\\
                $224\times224$ pixels}} \\
            Global caption supervision
                & \sharedcell{\topstack{1 short and 1 detailed caption per video,\\
                equal loss weights}} \\
            Global batch size
                & \sharedcell{4096} \\
            Optimizer
                & \sharedcell{\topstack{AdamW, $\beta_1=0.9$, $\beta_2=0.98$,\\
                weight decay $0.2$}} \\
            Text encoder learning rate
                & \sharedcell{$2\times10^{-5}$} \\
            Learning rate schedule
                & \sharedcell{\topstack{10 epochs total, 0.5 epoch of warmup,\\
                cosine decay}} \\
            Minimum learning rate multiplier
                & \sharedcell{$0.01$} \\
            Contrastive temperature $\tau$ (initial / minimum)
                & \sharedcell{$0.01$ / $0.01$} \\
            Vision drop path / dropout
                & \sharedcell{$0.1$ / $0$} \\
            Mixed precision
                & \sharedcell{BF16 autocast} \\
            \midrule
            Base learning rate
                & $2\times10^{-4}$ & $4\times10^{-5}$
                & $2\times10^{-4}$ & $4\times10^{-5}$ \\
            Joint embedding dimension
                & 512 & 768 & 512 & 768 \\
            \midrule
            Text context length (tokens)
                & \multicolumn{2}{c}{134}
                & \multicolumn{2}{c}{248} \\
            Position embedding stretch factor
                & \multicolumn{2}{c}{$2\times$}
                & \multicolumn{2}{c}{$4\times$} \\
            Visual patch masking
                & \multicolumn{2}{c}{50\%}
                & \multicolumn{2}{c}{10\%} \\
            Detailed caption fields in sentence bag
                & \multicolumn{2}{c}{--}
                & \multicolumn{2}{c}{4} \\
            Subcaptions per video / sentences per subcaption
                & \multicolumn{2}{c}{--}
                & \multicolumn{2}{c}{$12$ / $1$-$3$} \\
            Frame tokens per frame / registers per video
                & \multicolumn{2}{c}{--}
                & \multicolumn{2}{c}{1 / 2} \\
            Inter-frame attention layers
                & \multicolumn{2}{c}{--}
                & \multicolumn{2}{c}{2} \\
            \bottomrule
        \end{tabular}
    \end{subtable}

    \vspace{0.75em}

    \renewcommand{\sharedcell}[1]{%
        \multicolumn{5}{c}{\makebox[0pt][c]{#1}}%
    }

    \begin{subtable}{\textwidth}
        \centering
        \caption{Fine-tuning hyperparameters.}
        \label{tab:finetuning-hyperparameters}
        \begin{tabular}{@{}lccccc@{}}
            \toprule
            Hyperparameter & MSR-VTT & DiDeMo & ANet & MSVD & LSMDC \\
            \midrule
            Optimizer
                & \sharedcell{AdamW, $\beta_1=0.9$, $\beta_2=0.999$, weight decay $0.02$} \\
            Learning rate schedule
                & \sharedcell{1 warmup epoch, cosine decay to 1\% of the base learning rate} \\
            Global batch size
                & \sharedcell{256} \\
            Video input (train and test)
                & \sharedcell{12 frames at $224\times224$ pixels} \\
            Training augmentation
                & \sharedcell{Random resized crop (scale $0.5$--$1.0$) and horizontal flip} \\
            Text context length
                & \sharedcell{134 tokens (\baselinename), 248 tokens (\methodname)} \\
            \midrule
            Learning rate (both encoders)
                & $5\times10^{-6}$ & $5\times10^{-6}$ & $5\times10^{-6}$
                & $1\times10^{-5}$ & $5\times10^{-6}$ \\
            Training epochs
                & 7 & 8 & 5 & 5 & 10 \\
            Vision drop path rate
                & 0.3 & 0.2 & 0.3 & 0.2 & 0.3 \\
            \bottomrule
        \end{tabular}
    \end{subtable}
\end{table*}

\section{Training details}
\label{sec:training}
The hyperparameters used for model pretraining and fine-tuning are detailed in~\Cref{tab:hyperparameters,tab:finetuning-hyperparameters}, respectively. Notably, compared to \baselinename, \methodname increases the text context length from 134 to 248 tokens while reducing the visual patch masking ratio from 50\% to 10\%.

\section{Additional qualitative examples}
\label{sec:qual}
\begin{figure*}[p]
    \centering
    \begin{subfigure}[t]{\textwidth}
        \centering
        \includegraphics[width=\textwidth, height=0.43\textheight, keepaspectratio]{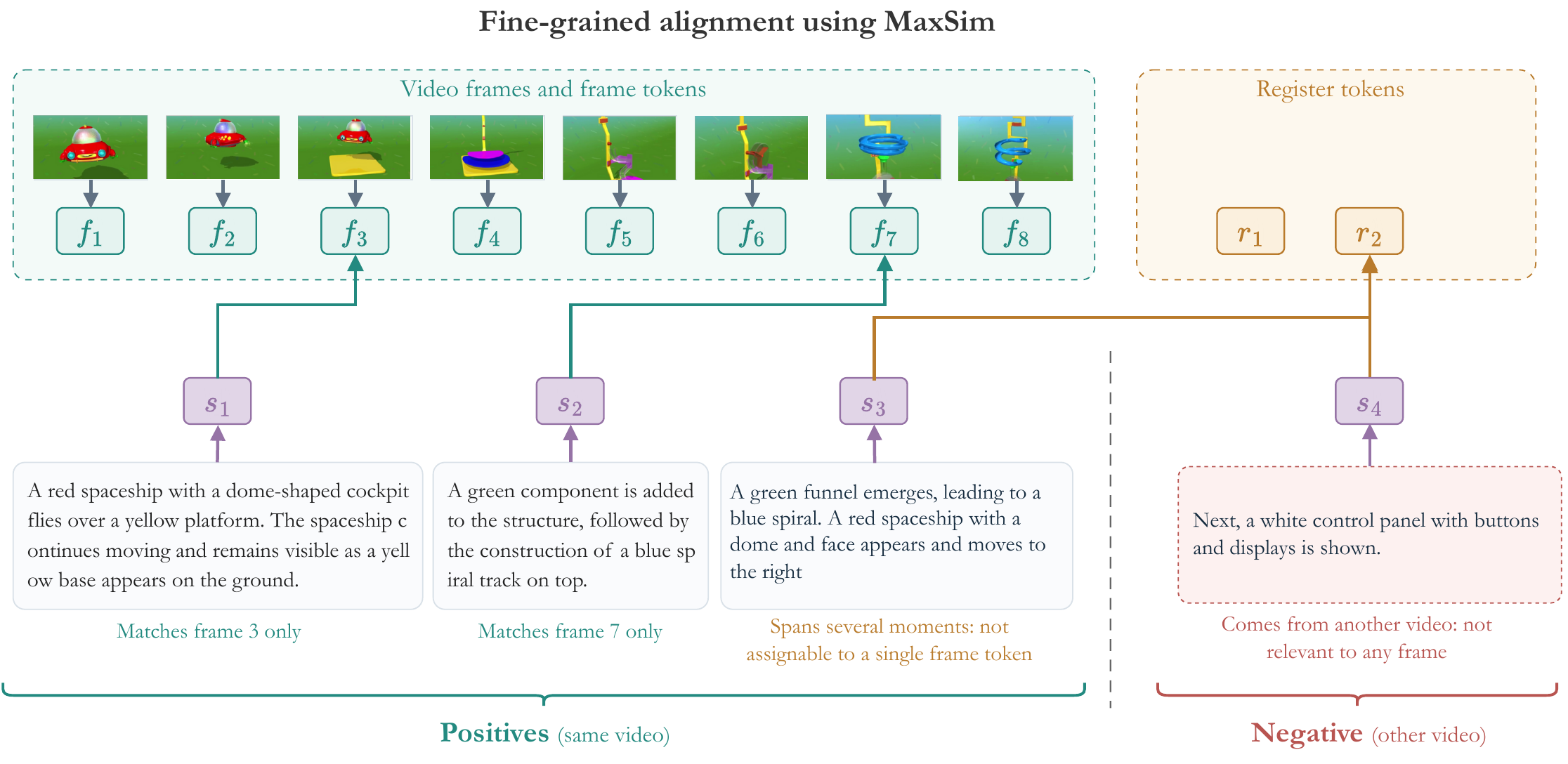}
        \caption{}
        \label{fig:example_a}
    \end{subfigure}

    \vfill

    \begin{subfigure}[t]{\textwidth}
        \centering
        \includegraphics[width=\textwidth, height=0.43\textheight, keepaspectratio]{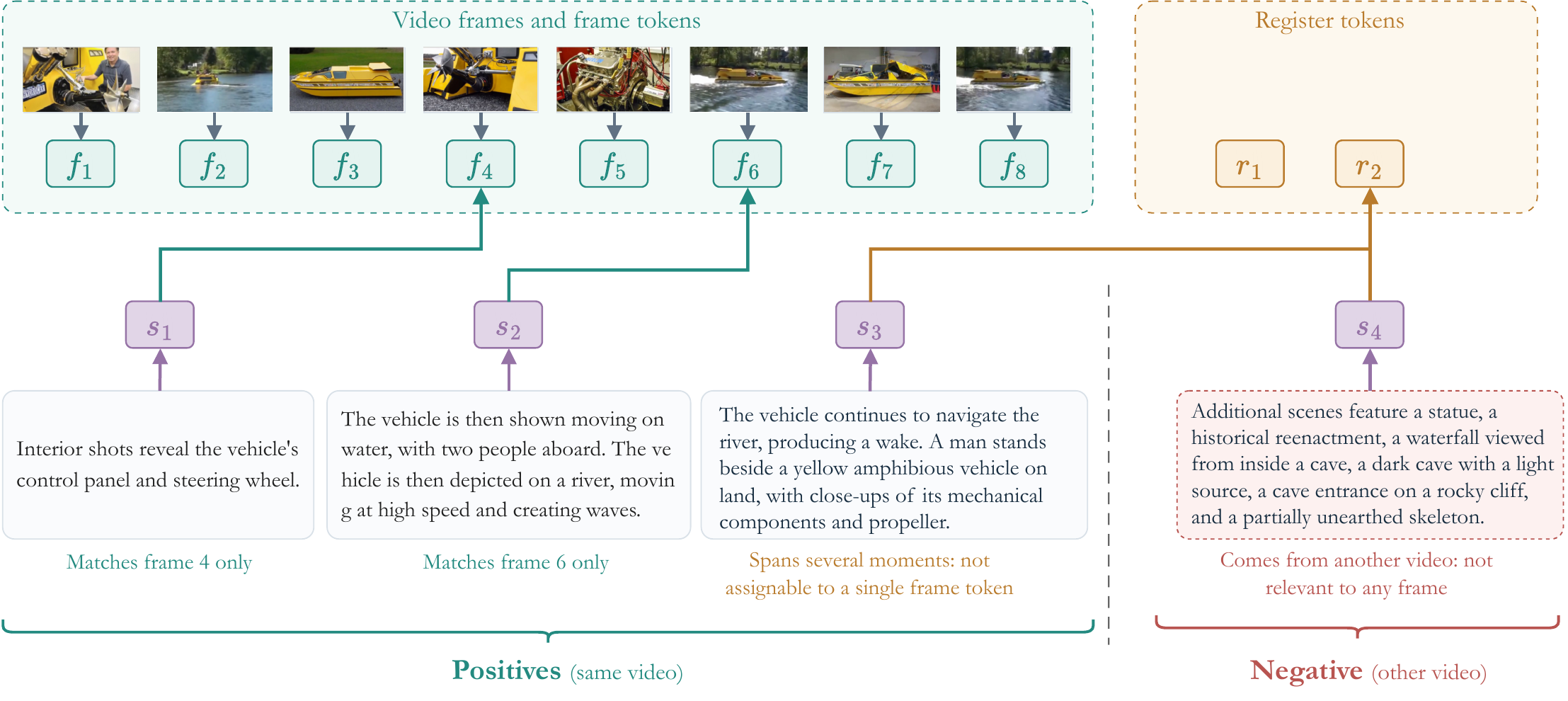}
        \caption{}
        \label{fig:example_b}
    \end{subfigure}

    \caption{\textbf{Additional qualitative examples of our MaxSim-based fine-grained alignment.}}
    \label{fig:example_suppl}
\end{figure*}

\begin{figure*}[p]
    \centering
    \begin{subfigure}[t]{\textwidth}
        \centering
        \includegraphics[width=\textwidth, height=0.29\textheight, keepaspectratio]{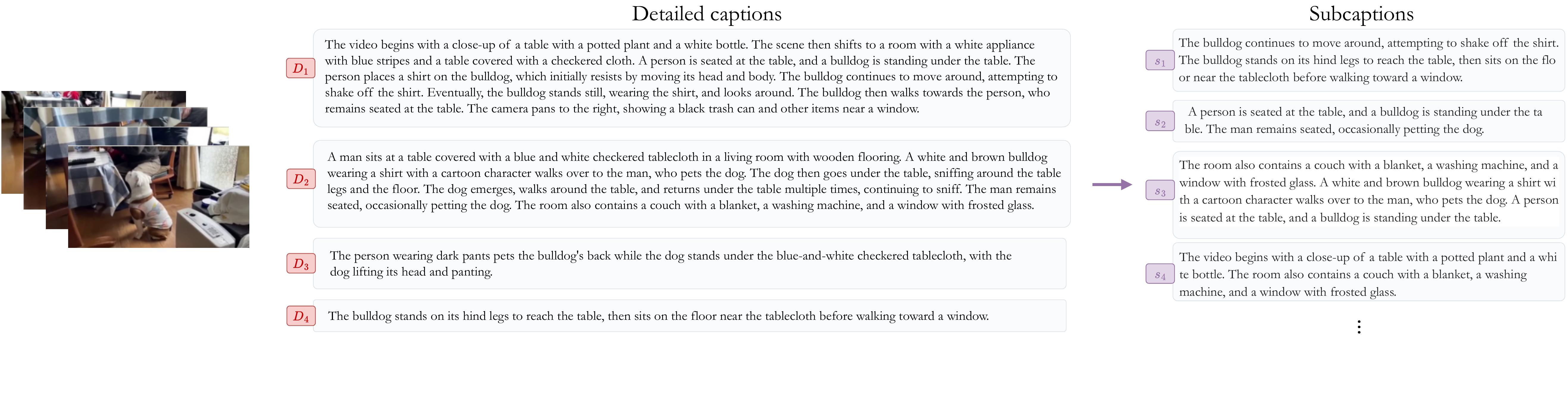}
        
        \label{fig:subcaption_example_a}
    \end{subfigure}

    \vfill

    \begin{subfigure}[t]{\textwidth}
        \centering
        \includegraphics[width=\textwidth, height=0.29\textheight, keepaspectratio]{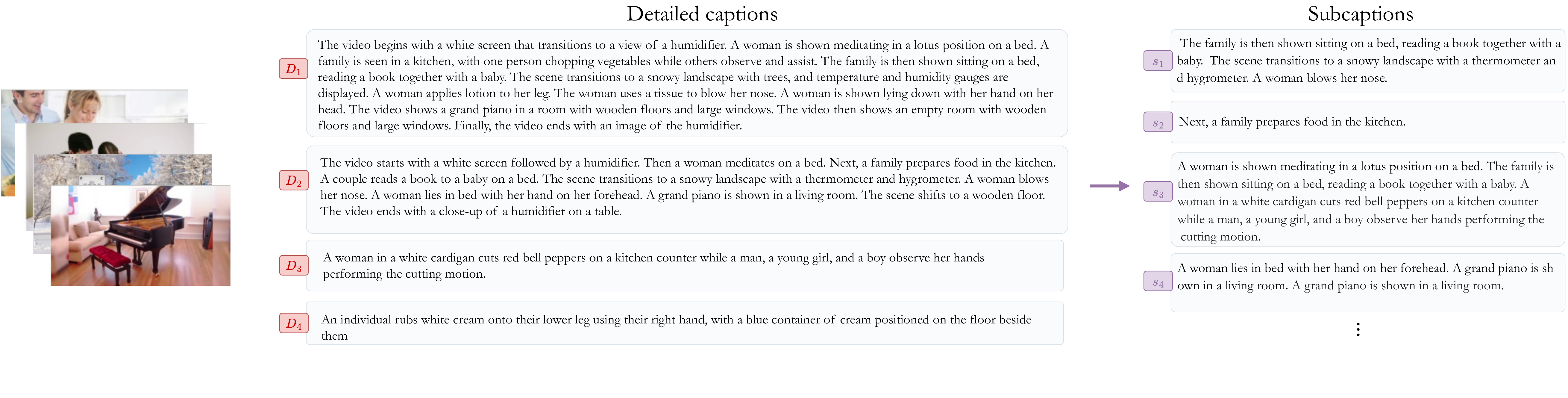}
        
        \label{fig:subcaption_example_b}
    \end{subfigure}

    \vfill

    \begin{subfigure}[t]{\textwidth}
        \centering
        \includegraphics[width=\textwidth, height=0.29\textheight, keepaspectratio]{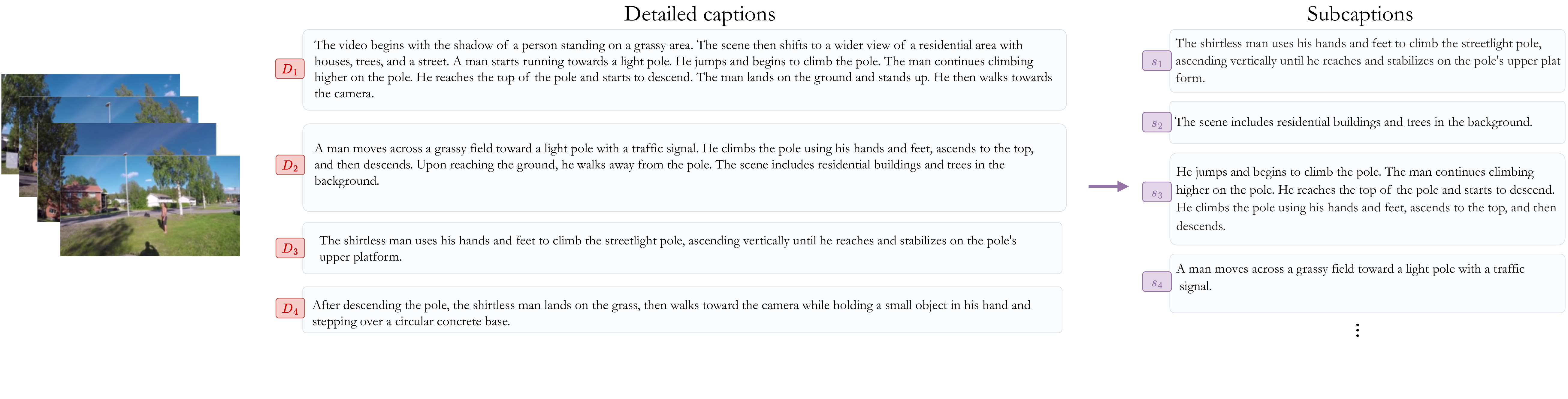}
        
        \label{fig:subcaption_example_c}
    \end{subfigure}

    \caption{\textbf{Illustration of our subcaption sampling strategy.}
    We split the detailed captions into individual sentences, randomly select between 1 and $N=3$ sentences, and concatenate them to form a subcaption. We repeat this process $K=12$ times for each video.}
    \label{fig:subcaption_example}
\end{figure*}

\begin{figure*}[p]
    \centering
    \begin{subfigure}[t]{\textwidth}
        \centering
        \includegraphics[width=\textwidth, height=0.43\textheight, keepaspectratio]{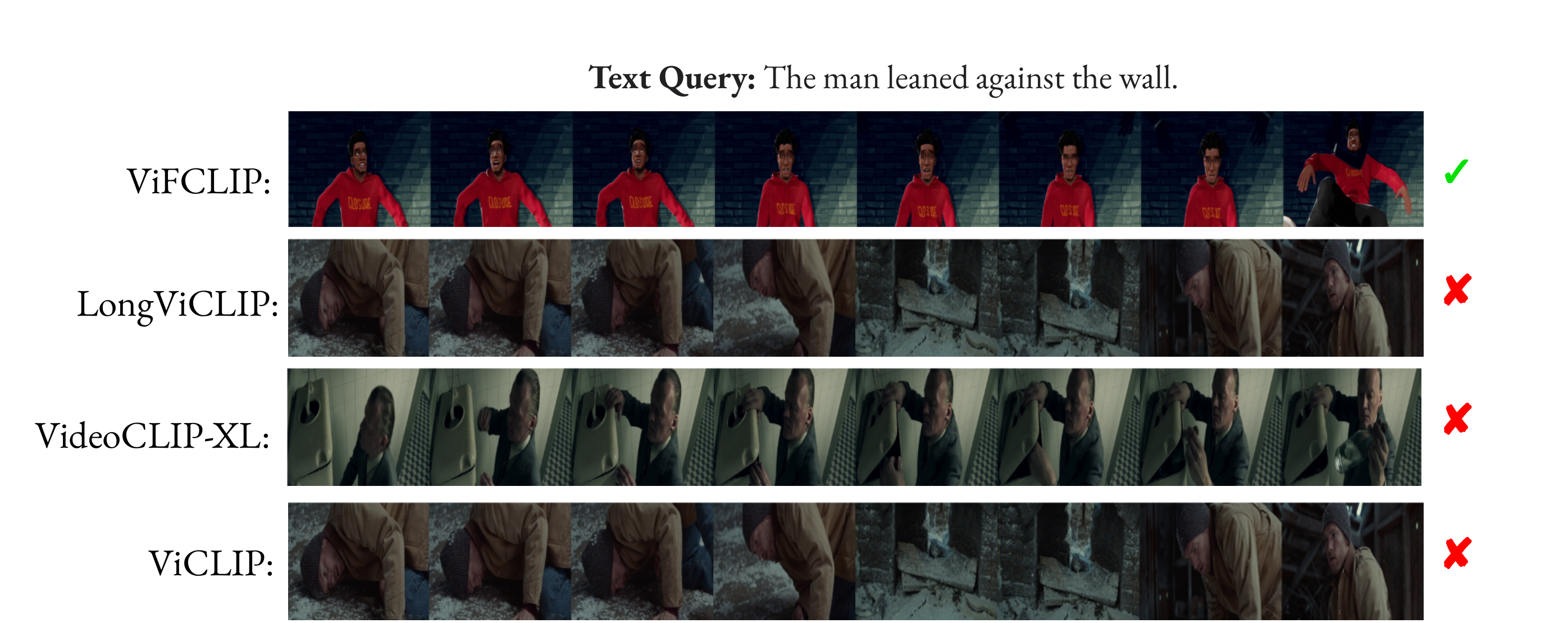}
        \caption{}
        \label{fig:retrieval_a}
    \end{subfigure}

    \vfill

    \begin{subfigure}[t]{\textwidth}
        \centering
        \includegraphics[width=\textwidth, height=0.43\textheight, keepaspectratio]{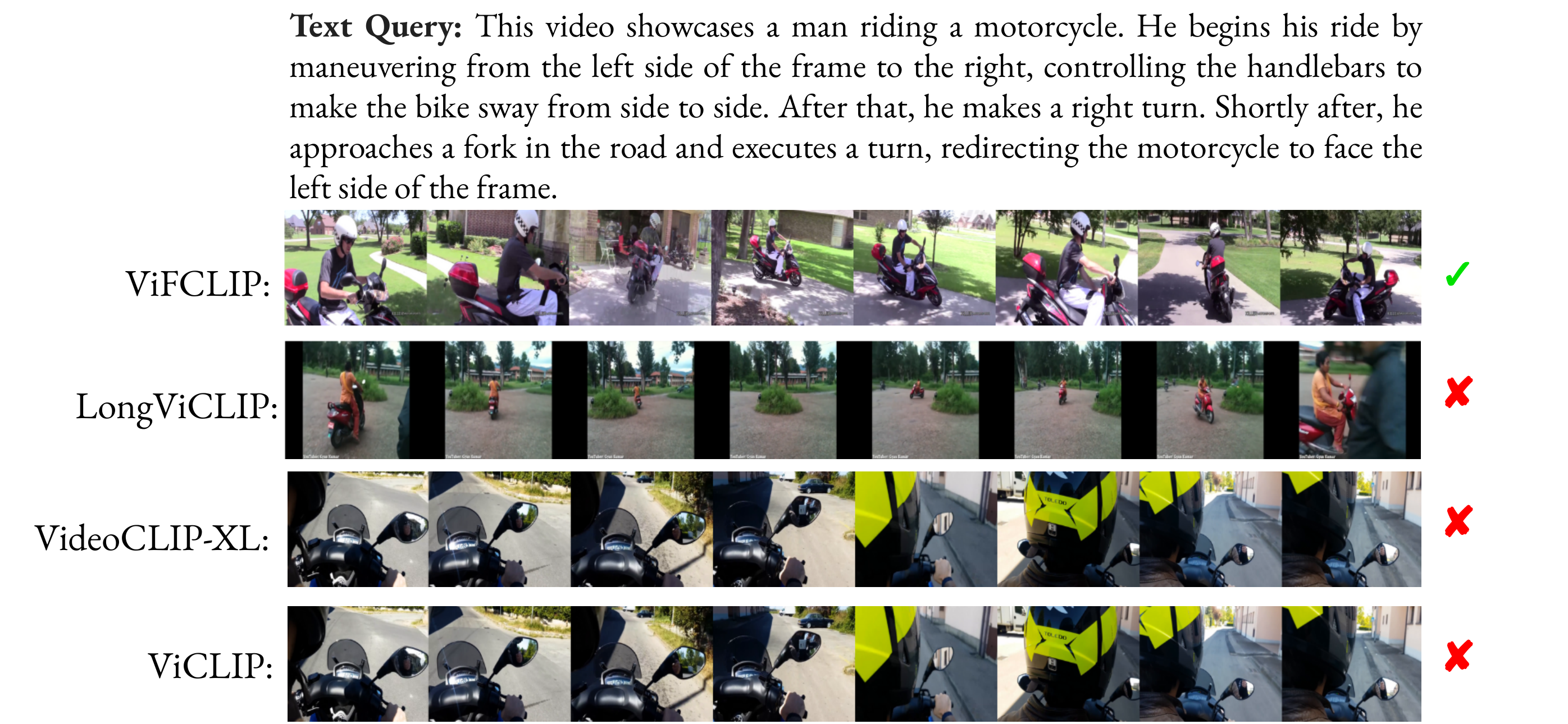}
        \caption{}
        \label{fig:retrieval_b}
    \end{subfigure}

    \caption{\textbf{Qualitative examples of fine-grained text-to-video retrieval.} (a) Example from DREAM-E. (b) Example from CRB-T.}
    \label{fig:retrieval_example}
\end{figure*}

~\Cref{fig:example_suppl} presents additional qualitative examples of the alignment between frame-tokens, register tokens and a subset of subcaptions. We also provide some examples of the subcaptions generated using our sampling strategy in ~\Cref{fig:subcaption_example}. Finally,~\Cref{fig:retrieval_example} shows qualitative examples of text-to-video retrieval on two fine-grained benchmarks CRB-T and DREAM-E. Compared to some CLIP-style models, \methodname better captures fine-grained details, leading to more accurate retrieval results.

\label{sec:qual-subcaption}

\label{sec:qual-retrieval}

\end{document}